\documentclass{article}

\usepackage{microtype}
\usepackage{graphicx}
\usepackage{subfigure}
\usepackage{booktabs} 
\usepackage{wrapfig}
\usepackage{multirow}
\usepackage{multicol}
\usepackage[backref=page]{hyperref}

\usepackage[accepted]{icml2024}

\usepackage{amsmath}
\usepackage{amssymb}
\usepackage{mathtools}
\usepackage{amsthm}
\usepackage[capitalize,noabbrev]{cleveref}
\theoremstyle{plain}
\newtheorem{theorem}{Theorem}[section]

\theoremstyle{definition}
\newtheorem{definition}[theorem]{Definition}
\newtheorem{assumption}[theorem]{Assumption}
\theoremstyle{remark}

\usepackage[most]{tcolorbox}
\usepackage{makecell}
\usepackage{bbding}
\usepackage{pifont}
\usepackage{tikz}

\usepackage[textsize=tiny]{todonotes}

\definecolor{mine}{RGB}{205, 232, 248}%

\icmltitlerunning{DecisionNCE: Embodied Multimodal Representations via Implicit Preference Learning}

\begin{document}

\twocolumn[
\icmltitle{
DecisionNCE: Embodied Multimodal Representations via \\ Implicit Preference Learning}
\icmlsetsymbol{equal}{*}

\begin{icmlauthorlist}
\icmlauthor{Jianxiong Li}{equal,thu}
\icmlauthor{Jinliang Zheng}{equal,thu,sensetime}
\icmlauthor{Yinan Zheng}{equal,thu}
\icmlauthor{Liyuan Mao}{thu,sjtu}
\icmlauthor{Xiao Hu}{thu}
\icmlauthor{Sijie Cheng}{thu}
\icmlauthor{Haoyi Niu}{thu}
\icmlauthor{Jihao Liu}{sensetime,mmlab}
\icmlauthor{Yu Liu}{sensetime}
\icmlauthor{Jingjing Liu}{thu}
\icmlauthor{Ya-Qin Zhang}{thu}
\icmlauthor{Xianyuan Zhan}{thu,shailab}
\end{icmlauthorlist}

\icmlaffiliation{thu}{AIR, Tsinghua University}
\icmlaffiliation{sensetime}{SenseTime Research}
\icmlaffiliation{sjtu}{Shanghai Jiaotong University}
\icmlaffiliation{mmlab}{CUHK MMLab}
\icmlaffiliation{shailab}{Shanghai AI Lab}

\icmlcorrespondingauthor{Jianxiong Li}{li-jx21@mails.tsinghua.edu.cn}
\icmlcorrespondingauthor{Xianyuan Zhan}{zhanxianyuan@air.tsinghua.edu.cn}

\icmlkeywords{Machine Learning, ICML}

\vskip 0.3in
]



\printAffiliationsAndNotice{\icmlEqualContribution} 

\begin{abstract}
Multimodal pretraining is an effective strategy for 
the trinity of goals of representation learning in autonomous robots: 
$1)$ extracting both local and global task progressions; $2)$ enforcing temporal consistency of visual representation; $3)$ capturing trajectory-level language grounding.
Most existing methods approach these via separate objectives, which often reach sub-optimal solutions.
In this paper, we propose a universal unified objective that can simultaneously extract meaningful task progression information from image sequences and seamlessly align them with language instructions. 
We discover that via implicit preferences, where a visual trajectory inherently aligns better with its corresponding language instruction than mismatched pairs, the popular Bradley-Terry model can transform into representation learning through proper reward reparameterizations. The resulted framework, \textit{DecisionNCE}, mirrors an InfoNCE-style objective but is distinctively tailored for decision-making tasks, providing an embodied representation learning framework that elegantly extracts both local and global task progression features, with temporal consistency enforced through implicit time contrastive learning, while ensuring trajectory-level instruction grounding via multimodal joint encoding.
Evaluation on both simulated and real robots demonstrates that DecisionNCE effectively facilitates diverse downstream policy learning tasks, offering a versatile solution for unified representation and reward learning. Project Page: \href{https://2toinf.github.io/DecisionNCE/}{https://2toinf.github.io/DecisionNCE/}
\end{abstract}
\section{Introduction}
\label{sec:intro}
Realizing general-purpose decision-making has long been an ultimate goal in AI research, which aims to build autonomous robotic agents that can comprehend environments via raw visual inputs and execute tasks described in natural language.
In pursuit of such intelligent agents, 
the brute-force solution is end-to-end training large models on extensive in-domain data with action annotations~\citep{rt2}. However, this can be exceedingly data-hungry, often posing a stratospheric demand for training data.
Pretraining scalable and generalizable vision-language representations, on the other hand, can leverage cheap out-of-domain data without action annotations, such as
in-the-wild human manipulation videos ~\citep{grauman2022ego4d}, to combat the scarcity of domain-specific data and facilitate effective downstream policy learning~\citep{r3m}. 
Although these generic videos may not directly subscribe to the specific downstream embodiment or tasks, their broad state coverage can greatly enhance generalizability. 
Moreover, these videos contain context-rich dynamics about how objects interact with their surroundings, and how task progressions are semantically inscribed in varied instructions~\citep{liv}. Hence, they hold immense potential for diverse downstream task executions, eliminating the dependency on intensive domain-specific data collection.

An imperative in learning effective multimodal joint representations is the \textit{trajectory-level grounding} of language, referring to aligning language with \textit{a sequence of images} rather than single frames, as image sequences (unlike static frames) inherently capture agent behaviors toward completing the given instructions~\citep{nair2022learning}.
In particular, 
focusing on image sequences of varied lengths is critical, as shorter sequences reveal detailed local transitions, while longer ones offer global task progression semantics. Striking the right balance between local and global views can be quite complex, yet 
it determines the overall quality of task representation~\citep{zhang2023closer}.
As shown in Table~\ref{tab:related_work}, existing methods tend to either focus on
short, fixed-length sequences, thus missing global context~\citep{karamcheti2023language, liv}; or consider only long horizons that overlook local transitions~\citep{r3m, nair2022learning}. Moreover, they often rely on separate objectives to either capture task progression, or enforce temporal consistency or language instruction alignment, delicately balanced by a hard-to-tune hyper-parameter~\citep{liv, r3m}, which often compromises certain objectives at the expense of others. Consequently, 
developing a principled solution that addresses these multifaceted challenges in a unified manner becomes crucial.

\begin{table}[t]
\small
\setlength{\tabcolsep}{1.1pt}
\caption{\small Comparison of vision-language representation learning methods for decision making. * Note that R3M can partially extract global and local progressions but is limited to always starting from the first image in a whole video, which overlooks intermediary transitions within a video. Please see Appendix~\ref{sec:related_works_appendix} for details.
}
\small
    \centering
    \resizebox{0.87\linewidth}{!}{
    \begin{tabular}{lcccccccc}
    \toprule
    Methods &\thead{Temporal\\consistency} &\thead{Global\\progressions} & \thead{Local\\progressions} &\thead{Unified\\learning}\\
    \midrule
    LOReL~(\citeyear{nair2022learning}) &\ding{55} &\Checkmark &\ding{55} &\ding{55}\\
    Voltron~(\citeyear{karamcheti2023language}) &\ding{55} &\ding{55} &\Checkmark &\ding{55} \\
    R3M~(\citeyear{r3m}) &\Checkmark &\Checkmark$^*$ &\Checkmark$^*$ &\ding{55} \\
    LIV~(\citeyear{liv}) &\Checkmark &\ding{55} &\Checkmark &\ding{55} \\
    \textbf{DecisionNCE} &\Checkmark &\Checkmark &\Checkmark &\Checkmark\\
    \bottomrule
    \end{tabular}
    }
    \label{tab:related_work}
\end{table}

\begin{figure}[t]
    \centering
    \resizebox{0.9\linewidth}{!}{
    \includegraphics[width=0.40\textwidth]{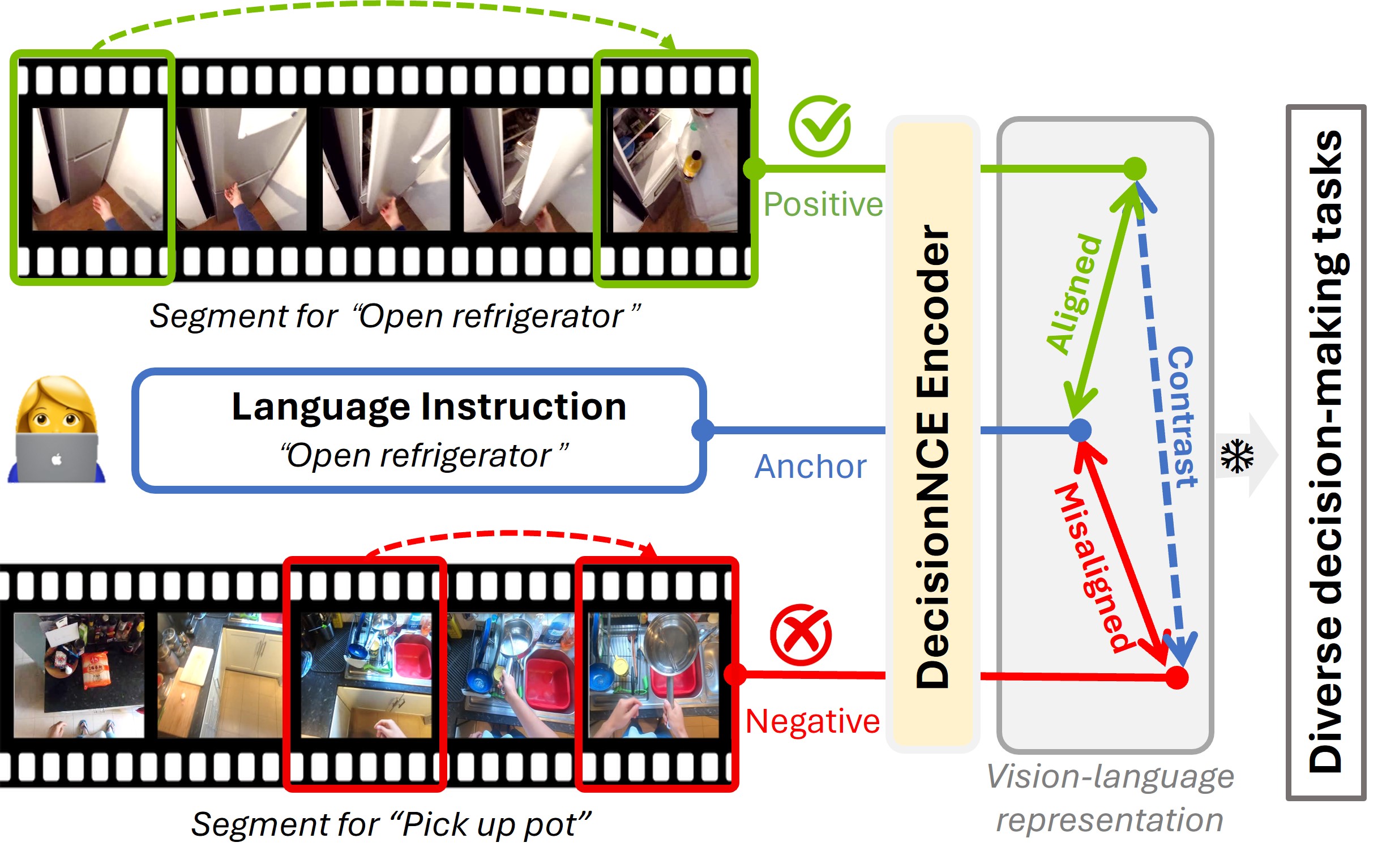}
    }
    \caption{\small \textbf{Implicit Preference Learning}: Matched segments and instructions are preferred to mismatches. Thus, implicit preference learning inherently performs a trajectory-level contrastive learning that compares segments rather than single images.}
    \label{fig:implicit_learning}
\end{figure}

To fill this gap, we introduce \textit{DecisionNCE}, a unified  multimodal representation learning framework for decision making.
DecisionNCE is realized via an elegant design that transforms 
the popular Bradley-Terry (BT) model~\citep{bradley1952rank} in preference-based reinforcement learning~\citep{akrour2011preference} into multimodal representation learning architecture. 
Specifically, we introduce the concept of \textit{implicit preferences}, where an image trajectory inherently aligns better with its corresponding language instruction than mismatched pairs  (illustrated in Figure~\ref{fig:implicit_learning}). Under this umbrella, BT model can transform into representation learning through proper reward reparameterizations. The resulted algorithm mirrors an InfoNCE-style~\citep{oord2018representation} objective, which is distinctively tailored for decision-making tasks, by shifting the contrast focus from single images to language-aligned trajectory segments.


Unlike previous work where instruction grounding is conducted on full video, which is expensive and time-consuming, 
we show that by introducing well-designed \textit{reward reparameterizations} in embedding spaces, the binary start-end transition in a selected video segment is sufficient in capturing task progression, bypassing irrelevant intermediary transitions to reduce computational cost and enable scalable learning. Specifically, we propose two DecisionNCE variants
: $1)$ DecisionNCE-P derives from the common inductive bias that final static images are more aligned with instructions~\citep{r3m, vip, liv}; and $2)$ DecisionNCE-T intuits that language instruction describes dynamical transition flows better than static images.
Notably, by adopting a random segment-sampling strategy, both DecisionNCE-P/T 
 simultaneously compare short and long segments of various lengths, from which local/global temporal information can be extracted over diverse time spans, which artfully mirrors implicit time contrastive learning~\citep{sermanet2018time}. This seamlessly fuses local/global task progression with temporal consistency, as well as language alignment in an elegant and unified objective, eliminating ad-hoc designs in previous works~\citep{r3m, vip, liv} and offering a surprisingly simple yet effective multimodal representation learning framework. 

Pretrained on large-scale, in-the-wild human video datasets, DecisionNCE excels in capturing temporally consistent task progression and comprehensively aligning with instructions. 
The learned representations from DecisionNCE can then be used for downstream policy learning. Extensive evaluations on both simulated and real robots demonstrate DecisionNCE's superiority to state-of-the-art representation learning methods. 
Moreover, the representations can also serve as zero-shot rewards parameterized in the embedding space, which can be used 
for zero-shot trajectory optimization. 

\section{Preliminaries}
\label{sec:preliminary}
\textbf{Notations}. 
We consider a language-conditioned Markov Decision Process (MDP) ~\citep{puterman2014markov} $\mathcal{M}:=\left( \mathcal{O}, \mathcal{A}, \mathcal{L}, r, \mathcal{P}, \gamma\right)$. $\mathcal{O}, \mathcal{A}$ are state and action space. $\mathcal{P}: \mathcal{O} \times \mathcal{A}\rightarrow\mathcal{O}$ is dynamic, and $\gamma \in (0,1)$ is discount factor. We assume the state space $\mathcal{O}=\mathbb{R}^{H\times W\times 3}$ is defined over RGB images. $\mathcal{L}$ is language instruction space, which specifies desired task behaviors. $r: \mathcal{O}\times\mathcal{A}\times\mathcal{L}\rightarrow\mathbb{R}$ is language-conditioned reward function.
The goal is to learn a policy $\pi: \mathcal{O}\times\mathcal{L}\rightarrow\mathcal{A}$ that can solve arbitrary tasks described in language $l\in\mathcal{L}$.

\begin{figure*}[t]
    \centering
    \includegraphics[width=0.95\textwidth]{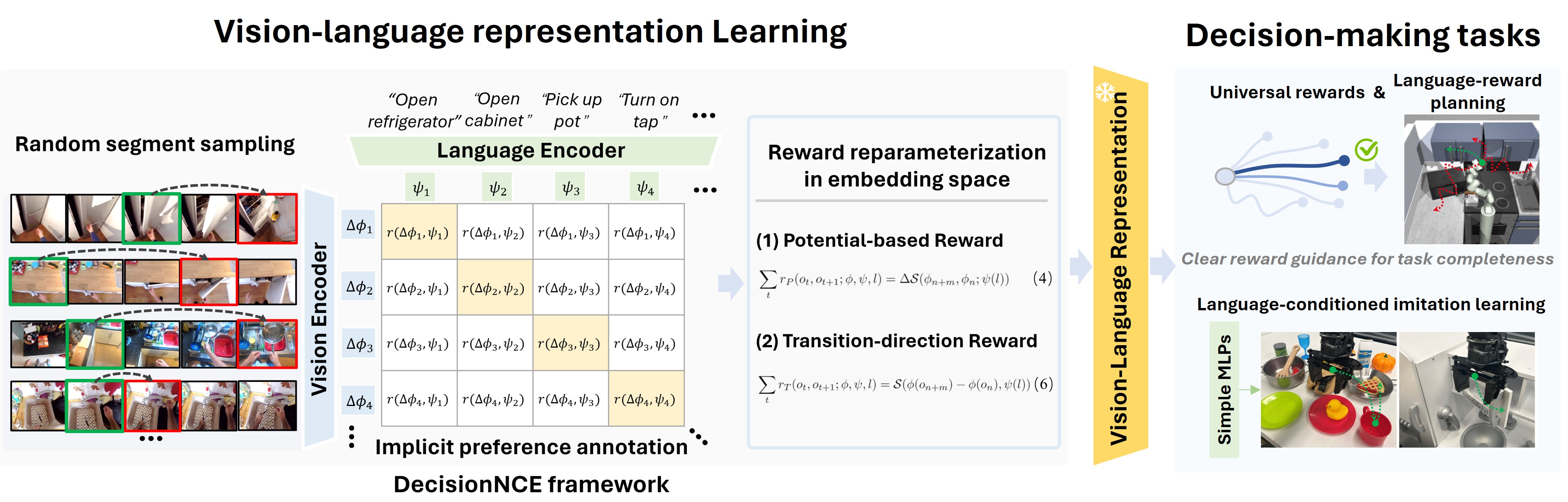}
    \caption{\small 
    Overview of DecisionNCE framework. DecisionNCE focuses on jointly training vision and language encoders to achieve trajectory-level representation alignment. The learned representations can be applied to various downstream decision-making tasks.}
    \label{fig:intro}
\end{figure*}
We assume a video dataset $\mathcal{D}=\{v_i:=(o_1^i, ..., o_{h_i}^i; l^i)\}_{i=1}^N$, where each $o\in \mathcal{O}$ is a raw RGB image, and each video $v$ comes with a descriptive language annotation $l$. $h$ is video length. The dataset $\mathcal{D}$ includes diverse out-of-domain data~\citep{grauman2022ego4d, damen2018scaling}, where no action labels are available. We aim to train a vision encoder $\phi: \mathbb{R}^{H\times W\times3}\rightarrow\mathbb{R}^K$ and a language encoder $\psi: \mathcal{L}\rightarrow\mathbb{R}^K$ to map the raw image $o$ and language $l$ into a shared $K$-dimensional vision-language representation space, using the dataset $\mathcal{D}$. After this, the policy $\pi(\phi(o), \psi(l))$ can leverage these representations as inputs for policy learning via imitation learning (IL) or reinforcement learning (RL).

\textbf{Bradley-Terry (BT) Model}. 
Designing proper rewards is quite challenging for complex tasks. In contrast, the well-known BT model~\citep{bradley1952rank} in preference learning
offers a natural way for humans to convey their desired outcomes to agents, where humans are required to rank pairs of segments. Then, BT model optimizes the rewards to 
maximize Eq.~(\ref{equ:bt_model_pre}) to fit human preferences~\citep{hu2023query}.
\begin{equation}
P\left[\sigma^+ \succ \sigma^-\right]=\frac{\exp \sum_{t} r\left(o_t^+, o_{t+1}^+\right)}{\sum_{i\in\{+, -\}}\exp \sum_{t} r\left(o_t^i, o_{t+1}^i\right)}.
\label{equ:bt_model_pre}
\end{equation}
where $\sigma=(o_n, ..., o_{n+m})$ denotes a length-$m$ segment. $\sigma^+\succ\sigma^-$ indicates segment $\sigma^+$ is preferred to $\sigma^-$, and the preference for $\sigma^+\succ\sigma^-$ grows exponentially with the total rewards of $\sigma^+$ and diminishes with those of $\sigma^-$.

\section{DecisionNCE}
\label{subsec:decisionNCE}
Note the BT model in Eq.~(\ref{equ:bt_model_pre}) is inherently a trajectory-level contrastive learning model that compares different trajectory segments and favors positive ones. 
This underlying equivalence inspires us to leverage BT model for representation learning. 
by reparameterizing the rewards with vision and language representations $\phi(o),\psi(l)$, and extending the original BT model into its language-conditioned version:
\begin{equation}
P\left[\sigma^+ \succ \sigma^-\right]=\frac{\exp \sum_{t} r\left(o_t^+, o_{t+1}^+;\phi,\psi,l^+\right)}{\sum_{i\in\{+, -\}}\exp \sum_{t} r\left(o_t^i, o_{t+1}^i;\phi,\psi,l^+\right)}.
\label{equ:bt_model}
\end{equation}
Eq.~(\ref{equ:bt_model}) essentially contrasts segments under various instructions,
to extract task-relevant semantic features from image sequence segments. 
This offers a new possibility for trajectory-level language grounding via adapting BT model. 

However, careful readers may notice that na\"ively extending Eq.~(\ref{equ:bt_model}) 
to trajectory-level grounding faces several challenges: \textit{1)} The preference labels are costly to collect~\citep{lee2021pebble}, and existing video-language datasets have no explicit preference annotations;
\textit{2)} It is tricky to select appropriate segment lengths $m$ for comparison, as overly long segments will induce extensive computations and lose local transition details, while short segments may lose global task information;
\textit{3)} It remains unclear what parameterized form of reward $r$ is best for multimodal representation learning.

To address these challenges, we introduce three novel designs: 
implicit preference annotations, random segment sampling, and reward reparametrization in embedding space. 

\subsection{Implicit Preference Annotations}
\begin{figure}[t]
    \centering
    \includegraphics[width=0.37\textwidth]{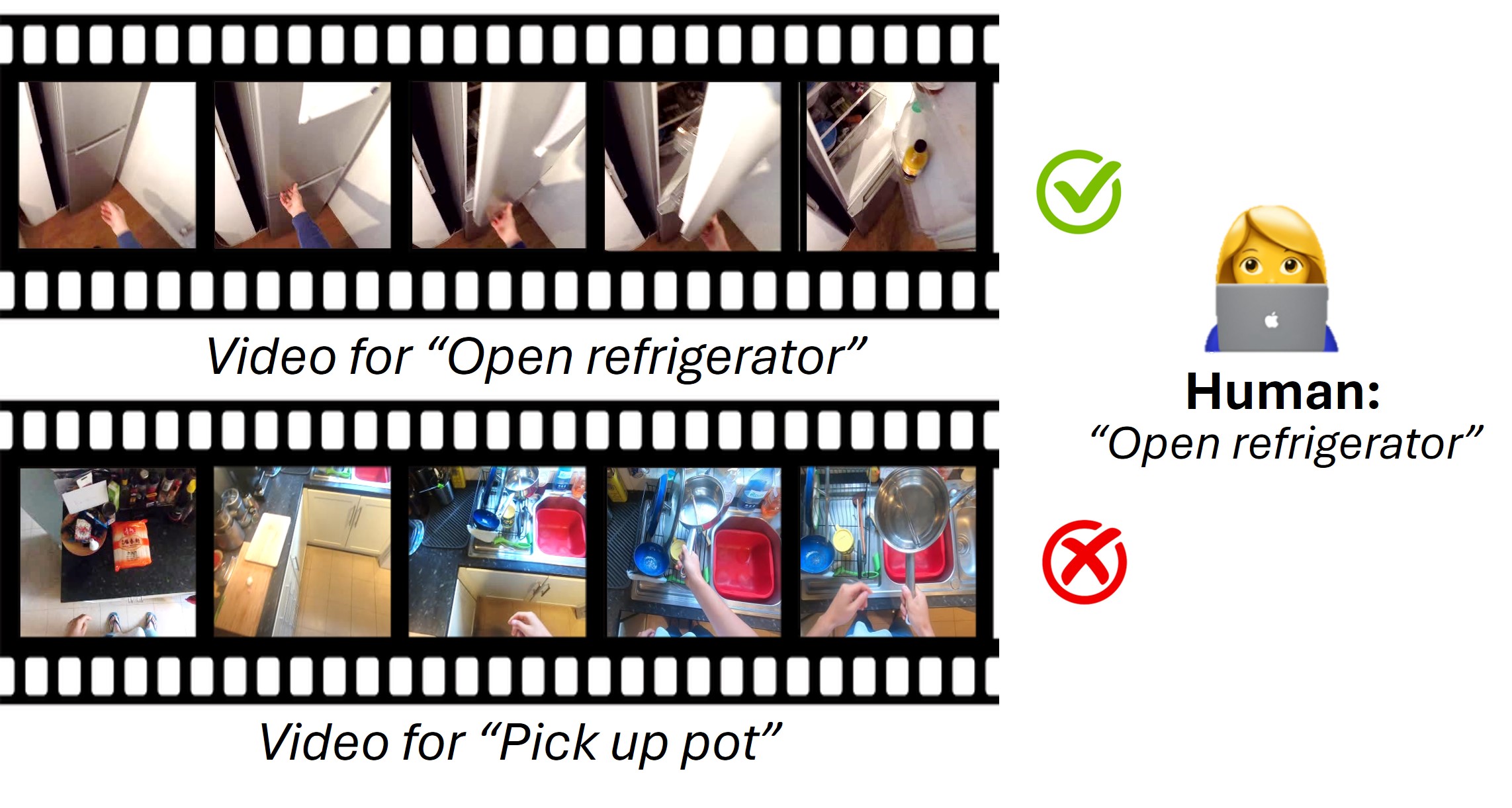}
    \caption{\small Implicit Preference. Segment is near-optimal for its associated language instruction, but is sub-optimal for others.}
    \label{fig:implicit_pref}
\end{figure}

Despite the absence of explicit preference labels in existing datasets, it's crucial to recognize that language instructions naturally communicate human intentions to agents. We can reasonably assume 
a video $v^i$ is more preferred for its associated language instruction $l^i$ than other mismatched instructions $l\neq l^i$, as shown in Figure~\ref{fig:implicit_pref}. This mild assumption holds in many cases. For example, a video on ``open microwave" is clearly sub-optimal for ``close microwave" instruction. Thus, the existing video-language dataset $\mathcal{D}$ already possesses a wealth of such \textit{implicit preference} that has yet to be fully exploited, eliminating the need for extra preference annotations. 

\subsection{Random Segment Sampling}
\label{subsec:radom_seg_sample}
Selecting a certain segment length $m$ that best covers both local and global information is challenging, and previous works typically focus on either short or long video segments~\citep{r3m, nair2022learning,liv, vip, karamcheti2023language} (see Appendix~\ref{sec:related_works_appendix_compare} for detailed discussion). 
A good balance between the two is preferable,
enjoying the complementary advantages of both~\citep{zhang2023closer}. To realize this, we adopt the widely used random-sampling strategy in action recognition for segment sampling~\citep{shi2013sampling, wang2018temporal, zhi2021mgsampler, zhang2023closer}. Specifically, we randomly select a start image $o_n$ from video $v=(o_1, ..., o_h)$ and then a goal image $o_{n+m}$ from the following frames $(o_{n+1}, ..., o_h)$. Repetitively, this leads to varied segment lengths $m$, enabling the preference model to both capture local transitions in short segments and global task progressions in longer ones. 

Despite the simplicity, we show in Section~\ref{subsec:analysis} that surprisingly, this simple sampling strategy induces implicit time contrastive learning for various DecisionNCE variants, guaranteeing smooth temporal consistency over time.

\subsection{Implicit Preference Learning via Reward Reparameterization}
\label{subsec:reward_para}
This section explains how to train the vision and language encoders $\phi(o), \psi(l)$. 
In principle, the DecisionNCE framework is versatile to any parameterized form of reward. Demonstratively, we show that with two simple reparameterization schemes in embedding space, i.e., \textit{potential-based} and \textit{transition-direction reward}, we can adeptly extend the original reward learning objective (Eq.~(\ref{equ:bt_model})) to effective representation learning
, allowing the optimization of vision and language encoder $\phi(o), \psi(l)$ to implicitly optimize rewards.

\textbf{(1) Potential-based Reward (DecisionNCE-P)}.  
Inspired by prior works that reformulate MDP in embedding space $\mathcal{M}_{\phi,\psi}:=\left( \phi(\mathcal{O}), \mathcal{A}, \psi(\mathcal{L}), r, \mathcal{P}, \gamma\right)$~\citep{li2022phasic, vip, liv}, we define the reward as the language-embedding distance difference in Eq.~(\ref{equ:r_def}):
\begin{equation}
\begin{aligned}
    &r_P(o_t, o_{t+1};\phi, \psi, l):=\mathcal{S}(\phi(o_{t+1}), \psi(l))-\mathcal{S}(\phi(o_{t}), \psi(l)). \\
\end{aligned}
\label{equ:r_def}
\end{equation}
Here, we choose cosine similarity for $\mathcal{S}$. This reward behaves similarly to potential-based reward shaping as $\mathcal{S}_{t+1}-\mathcal{S}_{t}=(1-\gamma)\mathcal{S}_{t+1} +(\gamma\mathcal{S}_{t+1}-\mathcal{S}_t)$, which is beneficial for policy learning~\citep{ng1999policy}.
Substitute Eq.~(\ref{equ:r_def}) into Eq.~(\ref{equ:bt_model}), the total reward for a segment $\sigma$ becomes:
\begin{equation}
    \sum_t r_P(o_t,o_{t+1};\phi,\psi,l)=\Delta\mathcal{S}(\phi_{n+m}, \phi_{n};\psi(l)),
    \label{equ:sum_r_def}
\end{equation}
where $\Delta\mathcal{S}(\phi_{n+m}, \phi_{n});\psi(l)):=\mathcal{S}(\phi(o_{n+m}), \psi(l))-\mathcal{S}(\phi(o_{n}),\psi(l))$ denotes the embedding distance shifts. Then, the resulting DecisionNCE-P instantiation is:
\begin{definition}[DecisionNCE-P]
\begin{equation}
P_P\left[\sigma^+ \succ \sigma^-\right]=\frac{\exp \Delta\mathcal{S}(\phi_{n+m}^+, \phi_{n}^+;\psi(l^+))}{\sum_{i\in\{+, -\}}\exp \Delta\mathcal{S}(\phi_{n+m}^i, \phi_{n}^i;\psi(l^+))}.
\label{equ:final_objective}
\end{equation}
\label{def:decision_P}
\end{definition}
Basically, maximizing Eq.~(\ref{equ:final_objective}) will attract $\phi(o_{n+m}^+)$ to $\psi(l^+)$ while repelling $\phi(o_{n}^+)$ from $\psi(l^+)$, leading to an inductive bias that \textit{final images are more aligned with instructions}~\citep{r3m, vip}.

\begin{figure}[t]
    \centering
    \includegraphics[width=0.38\textwidth]{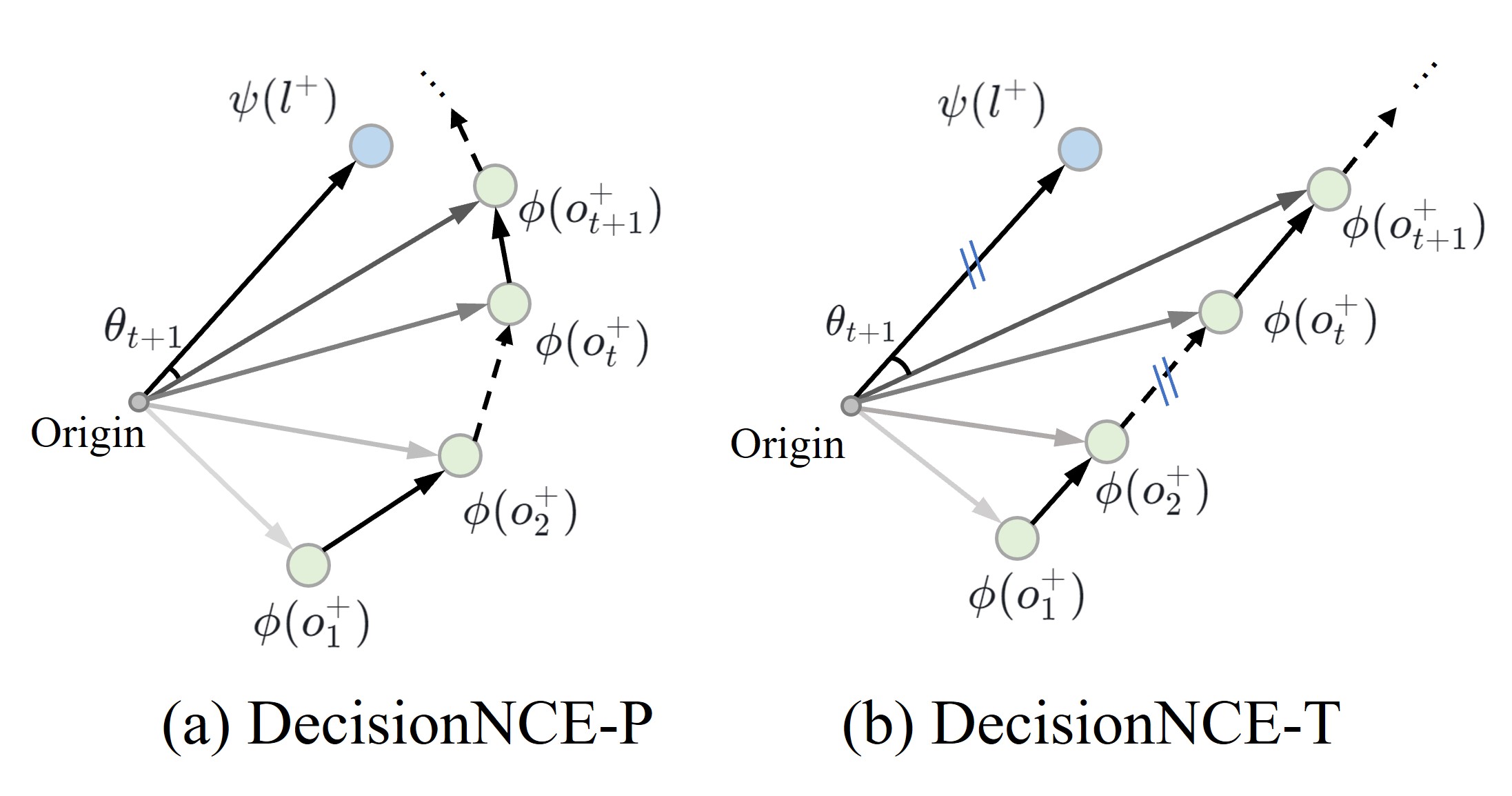}
    \caption{\small Illustration of DecisionNCE-P and DecisionNCE-T.}
    \label{fig:decisionNCE_pt}
\end{figure}

\textbf{(2) Transition-direction Reward (DecisionNCE-T)}. We also introduce an alternative inductive bias: \textit{language represents a transition direction instead of final images}, as instructions most likely describe dynamical behaviors rather than static images~\citep{karamcheti2023language}. To achieve this, we simply modify the total rewards in Eq.~(\ref{equ:sum_r_def}) to Eq.~(\ref{equ:sum_r_def_delta}), leading to the DecisionNCE-T variant in Definition~\ref{def:decision_T}.
\begin{equation}
    \sum_t r_T(o_t,o_{t+1};\phi,\psi,l)=\mathcal{S}(\phi(o_{n+m})-\phi(o_{n}),\psi(l)),
    \label{equ:sum_r_def_delta}
\end{equation}
\begin{definition}[DecisionNCE-T]
\begin{equation}
\small
P_T\left[\sigma^+ \succ \sigma^-\right]=\frac{\exp \mathcal{S}(\phi(o_{n+m}^+)-\phi(o_{n}^+),\psi(l^+))}{\sum_{i\in\{+, -\}}\exp \mathcal{S}(\phi(o_{n+m}^i)-\phi(o_{n}^i),\psi(l^+))}.
\label{equ:final_objective_T}
\end{equation}
\label{def:decision_T}
\end{definition}
Here, we obtain another representation learning objective. Maximizing Eq.~(\ref{equ:final_objective_T}) aligns the transition direction of $\phi(o_{n+m}^+)-\phi(o_{n}^+)$ and the text embedding vector $\psi(l^+)$. In this case, $\phi(o_{n+m}^+)$ and $\phi(o_n^+)$ can remain distant from $\psi(l^+)$ as long as their transition direction aligns effectively.

\begin{algorithm*}[t]
\footnotesize
\caption{DecisionNCE-P/T}
\begin{algorithmic}
    \STATE \textbf{Initialize:} Out-of-domain video dataset with language annotations $\mathcal{D}:=\{v_i:=(o_1^i, ..., o_{h_i}^i; l^i)\}_{i=1}^N$, vision-language encoder $\phi, \psi$
    \STATE \textbf{for} {each iteration} \textbf{do}:
        \STATE \quad \quad Randomly sample segments $\{\sigma_i\}_{i=1}^B=\{(o_{n_i}^i,...,o_{n_i+m_i}^i;l^i)\}_{i=1}^B$ from dataset $\mathcal{D}$ with random segment length $m_i$.
        \STATE \quad \quad $\mathcal{L}_{\rm DecisionNCE-P}(\phi, \psi)=\frac{1}{B}\sum_{i=1}^B\left[-\log\frac{\exp\Delta\mathcal{S}(\phi(o_{n_i+m_i}),\phi(o_{n_i});\psi(l^i))}{\sum_{j=1}^B\exp\Delta\mathcal{S}(\textcolor{blue}{\phi(o_{n_j+m_j}),\phi(o_{n_j}});\psi(l^i))}   - 
         \log\frac{\exp\Delta\mathcal{S}(\phi(o_{n_i+m_i}),\phi(o_{n_i});\psi(l^i))}{\sum_{j=1}^B\exp\Delta\mathcal{S}(\phi(o_{n_j+m_j}),\phi(o_{n_j});\textcolor{blue}{\psi(l^j)})}\right]$
         \STATE \quad \quad $\mathcal{L}_{\rm DecisionNCE-T}(\phi, \psi)=\frac{1}{B}\sum_{i=1}^B\left[-\log\frac{\exp\mathcal{S}(\phi(o_{n_i+m_i})-\phi(o_{n_i}), \psi(l^i))}{\sum_{j=1}^B\exp\mathcal{S}(\textcolor{blue}{\phi(o_{n_j+m_j})-\phi(o_{n_j})}),\psi(l^i))}   - 
         \log\frac{\exp\mathcal{S}(\phi(o_{n_i+m_i})-\phi(o_{n_i}),\psi(l^i))}{\sum_{j=1}^B\exp\mathcal{S}(\phi(o_{n_i+m_i})-\phi(o_{n_i}),\textcolor{blue}{\psi(l^j)})}\right]$
         \STATE \quad \quad Update $\phi$ and $\psi$ using SGD: $(\phi, \psi)\leftarrow(\phi, \psi)-\alpha\nabla\mathcal{L}_{\rm DecisionNCE-P}(\phi, \psi)$ or $(\phi, \psi)\leftarrow(\phi, \psi)-\alpha\nabla\mathcal{L}_{\rm DecisionNCE-T}(\phi, \psi)$
\end{algorithmic}
\label{algo:decisionNCE}
\end{algorithm*}

\textbf{Connections between DecisionNCE-P/T}. Comparing the two variants, we notice that DecisionNCE-P explicitly concentrates on the absolute similarities between single images and text instructions, whereas DecisionNCE-T is less concerned with this aspect, but emphasizes more on the correctness of the relative transition direction. However, DecisionNCE-T can implicitly ensure a correct cosine similarity between $\phi(o_{t+1}^+)$, $\phi(o_t^+)$, and $\psi(l^+)$. As shown in Figure~\ref{fig:decisionNCE_pt} (b), their angle $\theta_t$ can monotonically decrease if the transition direction is correct. Thus, Decision-T not only explicitly enforces a meaningful embedding learning direction, but also implicitly guarantees correct absolute embedding positions like DecisionNCE-P.

\subsection{Practical Implementation}
The pseudo-code of DecisionNCE for practical implementation is presented in Algorithm~\ref{algo:decisionNCE}. Specifically, we extend the single negative segment in Eq.~(\ref{equ:final_objective}),~(\ref{equ:final_objective_T}) to all mismatched pairs that appeared in a mini-batch to improve training efficiency (highlighted as blue), as one gradient step can establish contrastive learning signals to more negative samples. This does not affect the analysis in Section~\ref{subsec:analysis}, since the optimization direction remains unchanged for positive and negative segments. This final objective resembles an InfoNCE-style~\citep{oord2018representation} loss function, but the contrastive terms are start-end transitions rather than single images. Thus, we term our method as \textit{DecisionNCE}. Please see Appendix~\ref{appen:exp_details} for more implementation details.

\section{Analyses and Insights}
\label{subsec:analysis}
Given the two instantiations of DecisionNCE, in the following, we offer more in-depth analyses and insights of the algorithm designs and potentials of the proposed framework.
\subsection{From Full Segments to 
Start-end Transitions}
\label{subsubsec:full-seg-to-start-end}

Note that by reparameterizing the reward as in Eq.~(\ref{equ:sum_r_def}), the intermediary transitions are conveniently canceled out. This greatly simplifies the comparison of full segments $\sigma$ to just a binary start-goal transition $o_n\rightarrow o_{n+m}$ in any segment, 
which has proven 
advantageous in action 
recognition~\citep{korbar2019scsampler, zhang2023closer}.
\begin{wrapfigure}{r}{3.45cm}
\hspace{-10pt}
    \centering
    \includegraphics[width=0.19\textwidth]{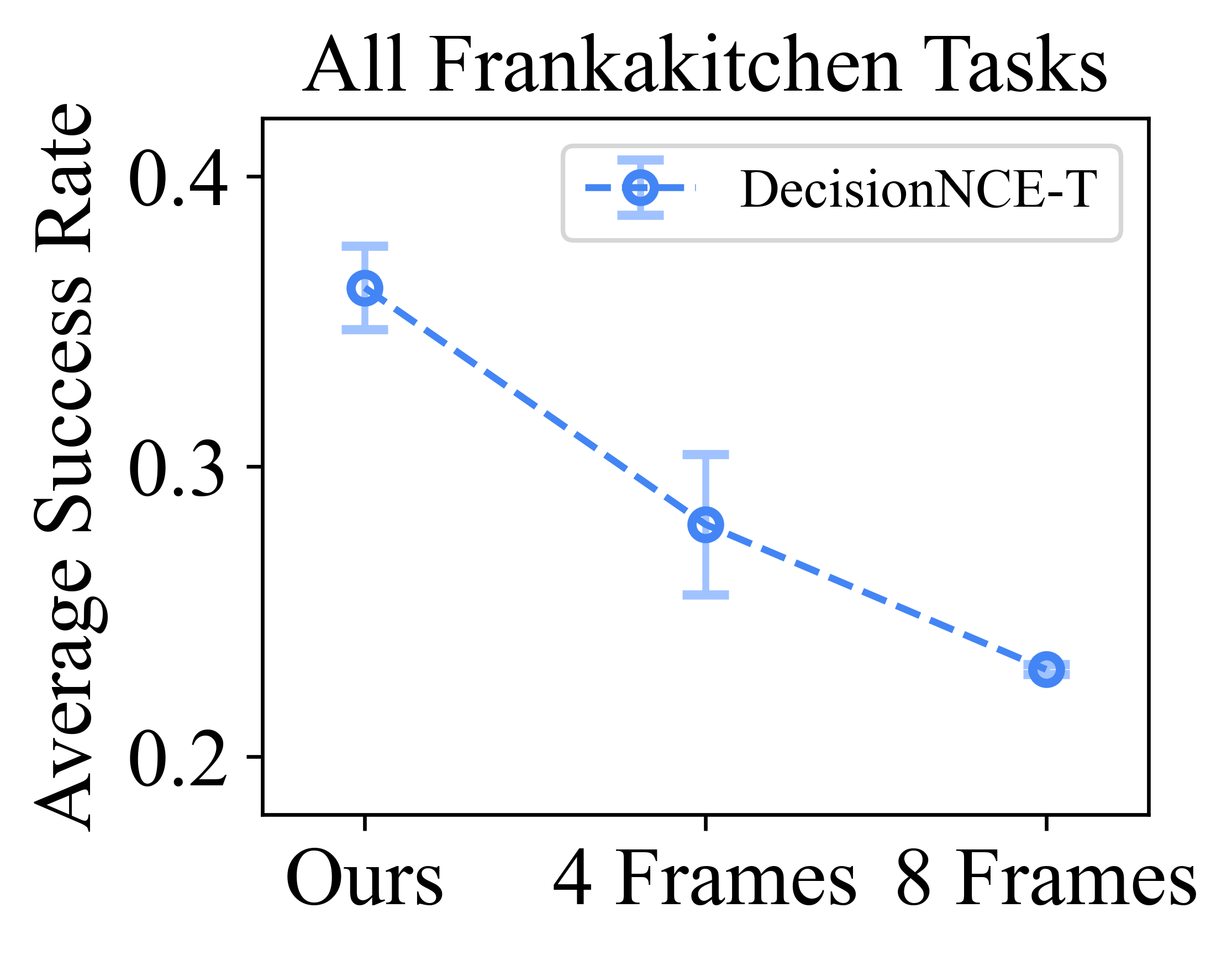}
    \caption{\small Ablation on different numbers of frames used for preference learning. }
    \label{fig:ablation_248}
\end{wrapfigure}
This can effortlessly scale to any segment length, and by doing so, training efficiency is significantly improved, as comparing every transition 
within segments can be prohibitively expensive. Moreover, comparing all transitions often covers numerous meaningless intermediary points, especially over long segments, which can overwhelm informative transitions with noise and complicate training. 
In contrast, Eq.~(\ref{equ:sum_r_def}) concentrates on the critical transition $o_n\rightarrow o_{n+m}$, maintaining the essential task progression while bypassing most noisy data. To investigate this, in Figure~\ref{fig:ablation_248}, we train DeicisionNCE-T using the total segment rewards calculated by not only $o_n\rightarrow o_{n+m}$, but also including some intermediary points (see Appendix~\ref{subsec:appen_248} for experiment setups). The results show that including intermediary transitions sometimes can be harmful for representation learning, leading to decreased success rate on policy learning. Therefore, we design our \textit{transition-direction reward} in Eq.~(\ref{equ:sum_r_def_delta}) solely on the start-goal transitions.

\subsection{Mirroring Time Contrastive Learning} At first glance, some works also compare binary transitions 
similar to DecisionNCE-P/T, but they are limited to either long transitions $o_1\rightarrow o_h$ across entire trajectories~\citep{nair2022learning} or short transitions with a fixed $k$-interval $o_t\rightarrow o_{t+k}$~\citep{liv, vip, karamcheti2023language}, and require additional objective for language grounding, which leads to complicated trade-offs among various elements. By contrast, we find that our segment comparisons in Eq.~(\ref{equ:final_objective}) and Eq.~(\ref{equ:final_objective_T}) effortlessly integrate implicit time contrastive learning for temporal consistency and meanwhile conduct local/global language grounding within a single, cohesive objective, effectively overcoming these challenges. 

\begin{figure*}[t]
    \centering
    \includegraphics[width=0.98\textwidth]{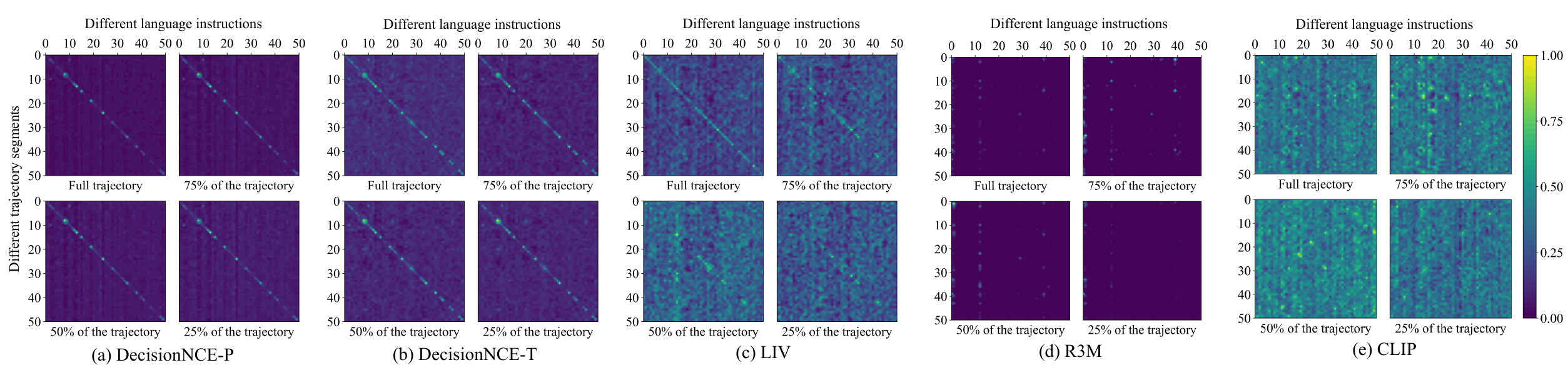}
    \caption{\small Heatmap of the learned rewards for various instruction/segment pairs. The diagonals are matched pairs and off-diagonals denote mismatches. DecisionNCE identifies unmatched pairs with varied segment lengths, extracting both local/global task progressions.}
    \label{fig:heatmap}
\end{figure*}

Specifically, a segment $\sigma^+$ is randomly selected from $v^+$, so any frame in $v^+$ can be a ``goal" image or a ``start" image. However, thanks to the \textit{random segment sampling} strategy, the chance for frame $o_t^+$ being selected as the goal is not isotropic but gradually increases as the video progresses. Mathematically, the probability for $o_t^+, 0<t<h$ being a goal image grows monotonically w.r.t its time stamp $t$ (see Appendix~\ref{appen:proof} for proof):
\begin{equation}
\vspace{-3pt}
\begin{aligned}
    &P(o_t^+ {\rm \ is\ selected\ as\ goal})=\frac{1}{h}\sum_{i=1}^{i<t}\frac{1}{h-i}
    \label{equ:strength}
\end{aligned}
\end{equation}

Therefore, for DecisionNCE-P, Eq.~(\ref{equ:strength}) clearly implies that later images $o_j^+$ have an increasing likelihood of being attracted to $\psi(l^+)$ than earlier ones $o_i^+, j>i$, with the attraction strength smoothly increasing over time step $t$. Thus, DecisionNCE-P implicitly promotes smooth temporal progressions in representation spaces, mirroring time contrastive learning~\citep{sermanet2018time}. This characteristic also holds for DecisionNCE-T. As depicted in Figure~\ref{fig:decisionNCE_pt}(b), as the transition directions between randomly sampled image embedding pairs are driven to be similar within the same video, the cosine similarity between consecutive image embeddings $\phi(o_t), \phi(o_{t+1})$ can also remain large.
Indicated by Eq.~(\ref{equ:strength}), the transition direction from initial to later image embeddings will be increasingly aligned with the direction of text embedding vector, as later images are more frequently chosen as goal images in video trajectories.

Notably, both DecisionNCE-P/T seamlessly integrate language grounding and temporal consistency into a single, simplified loss function, avoiding the intricate balance required between these two objectives. Additionally, the use of randomly selected segments eliminates the need for extra tedious hyper-parameter tuning, unlike prior methods that rely on fixed or meticulously adjusted intervals~\citep{vip, liv, karamcheti2023language, r3m, nair2022learning}.

\subsection{Positioning Task-Irrelevant Image Embeddings}
In DecisionNCE-P/T,  frames in negative segments $\sigma^-=(o_n^-, ..., o_{n+m}^-;l^-)\subseteq v^-=(o_1^-,...,o_h^-;l^-)$ are optimized oppositely as compared to positive segments. This can potentially attract sampled initial image embedding $\phi(o_{n}^-)$ to $\psi(l^+)$, and the first image $o_1^-$ in negative video $v^-$ will be the most attracted to $\psi(l^+)$. Counter-intuitively, this seems ``incorrect" at first glance, as $o_1^-$ is typically irrelevant to positive language instructions. However, we can demonstrate that this seemingly ``incorrect" objective will actually lead to desired representation spaces, unlike other methods. 

\begin{figure*}[t]
    \centering
    \includegraphics[width=0.99\textwidth]{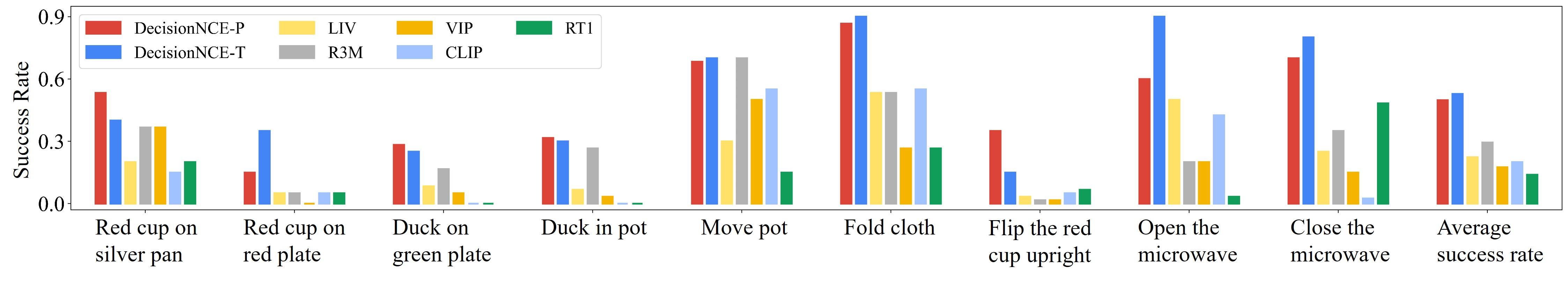}
    \vspace{-5pt}
    \caption{\small \textbf{Real robot LCBC experimental results}. Success rate is averaged over 10 episodes and 3 seeds.}
    \label{fig:realrobot}
\end{figure*}

\begin{table}[t]
\small
\setlength{\tabcolsep}{0.8pt}
\caption{\small Cosine similarities among the first image embeddings $\phi(o_1)$ and similarities to language embeding means $\psi(\tilde{l})$. See Appendix~\ref{subsec:appen_table2} for experimental setups.}
    \centering
    \resizebox{1.0\linewidth}{!}{
    \begin{tabular}{lccc}
    \toprule
    \multirow{2}{*}~  &\multirow{2}{*}{LIV} &\multicolumn{2}{c}{DeicisionNCE (Ours)}\\
    \cmidrule(r){3-4}
    ~ & ~ & -P & -T \\
    \midrule
    $\phi(o_1)$ similarities$\uparrow$  &$0.08 \pm 0.005$ &\quad \colorbox{mine}{$0.44 \pm 0.01$} &\quad \colorbox{mine}{$0.96 \pm 0.004$}\\
    Similarities to $\psi(\tilde{l})\uparrow$  &$-0.06 \pm 0.02$ &\quad \colorbox{mine}{$0.04 \pm 0.008$} &\quad \colorbox{mine}{$0.02 \pm 0.01$} \\
    \bottomrule
    \end{tabular}}
    \label{tab:cosine_sim_o1}
\end{table}

\begin{wrapfigure}{r}{3.3cm}
    \centering
    \includegraphics[width=0.2\textwidth]{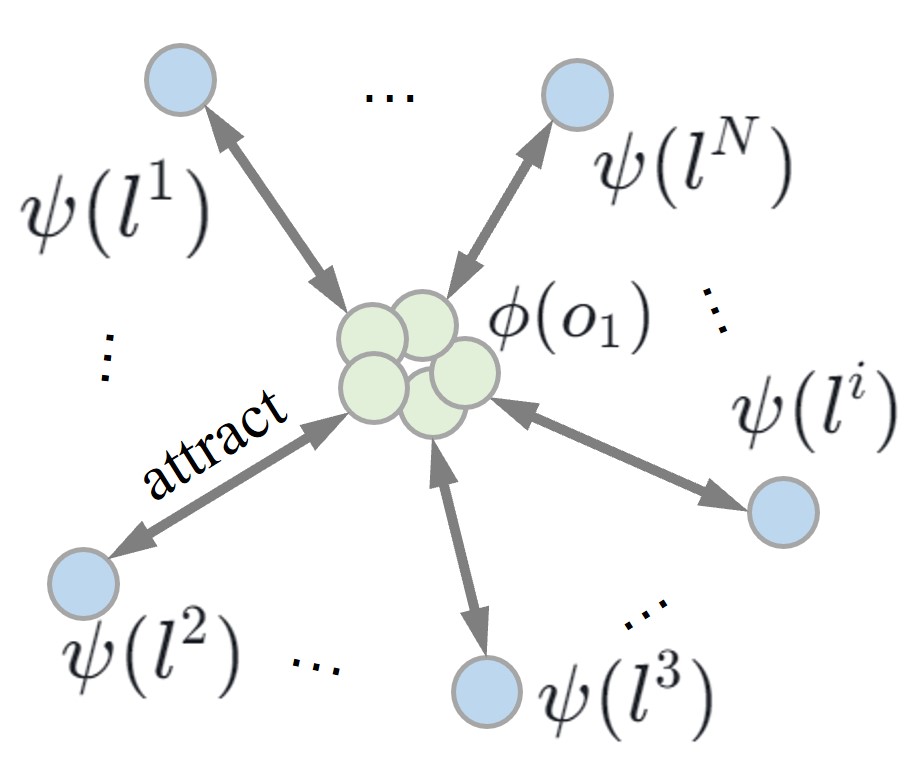}
    \vspace{-20pt}
    \caption{\small Illustration for task-irrelevant first image embeddings $\phi(o_1)$.}
    \label{fig:phi_o1}
\end{wrapfigure}
In fact, based on our designs in DecisionNCE,
$\phi(o_1^-)$ is not only drawn to $\psi(l^+)$, but to all mismatched language embeddings with $l\neq l^-$. This essentially drives first image embeddings $\phi(o_1)$ of different videos to the central position among all language embeddings,
as depicted in Figure~\ref{fig:phi_o1}. 
This is actually desirable, as 
the first images of videos hardly carry any substantial information about the goal of tasks.
Determining the nature of the tasks based solely on the first images in videos is highly difficult even for humans. As an example,
we notice that in video-language datasets like KITCHEN-100~\citep{damen2018scaling}, the videos for ``open door" instruction might not even show a door in the first image.
So ideally, $\phi(o_1)$ should be positioned away from any specific instruction as it lacks task-specific information, which is exactly what 
DecisionNCE-P/T do. In  Table~\ref{tab:cosine_sim_o1}, we demonstrate that DecisionNCE-P/T successfully cluster potentially task-irrelevant first image embeddings into similar positions, yielding high similarities among them, whereas other methods cannot.

\subsection{Advanced Local/Global Trajectory-level Grounding}
Beyond previously discussed properties, we also examine the ability of DecisionNCE to capture both local and global task progressions. Specifically, similarities for segments of varied lengths should be higher for matched instructions than mismatched pairs. The visualization in Figure~\ref{fig:heatmap} clearly demonstrates that DecisionNCE is superior in this aspect, as our training strategy explicitly guarantees this with the effective \textit{random segment sampling} scheme under the implicit preference learning framework, capturing local and global task progressions by comparing both short and long segments.
Previous methods, however, are often impaired by noises.

\section{Experiments}
\label{sec:experiments}

\begin{figure}[h]
    \centering
    \includegraphics[width=0.44\textwidth]{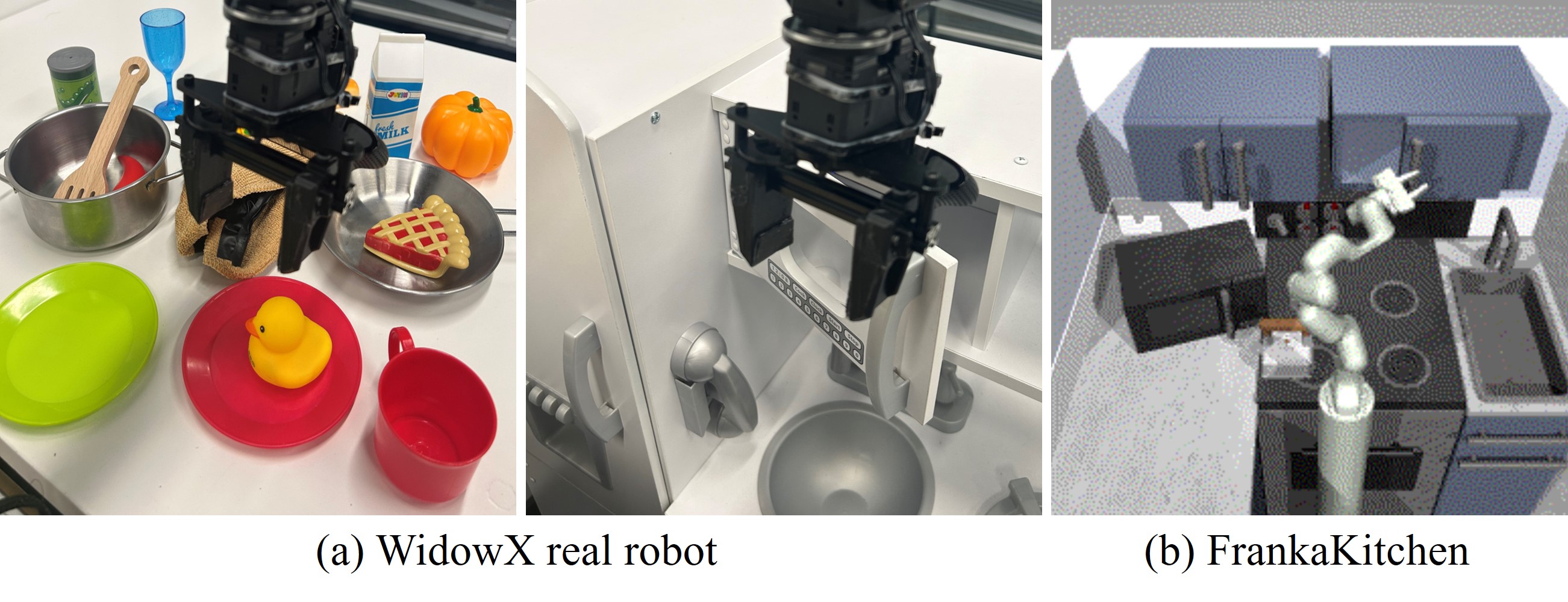}
     \vspace{-10pt}
    \caption{\small Experimental environments.}
    \label{fig:env}
\end{figure}

We pretrain the DecisionNCE encoders using large-scale human video dataset EPIC-KITCHEN-100~\citep{damen2018scaling} and conduct extensive experiments on both simulated and real robotic environments (Figure~\ref{fig:env}). Specifically, we investigate whether the representation learned by DecisionNCE can power effective downstream policy learning using methods like Language-Conditioned Behavior Cloning (LCBC), and whether DecisionNCE can offer universal language-conditioned rewards that describe the desired task progression directions.

\textbf{Baselines.} We compare DecisionNCE with following baselines: $1)$ \textit{CLIP} \citep{radford2021learning}: Aligns images and language through contrastive learning. $2)$ \textit{R3M} \citep{r3m}: Combines LOReL~\citep{nair2022learning} and time contrastive learning~\citep{sermanet2018time} with a frozen language encoder. 
$3)$ \textit{VIP} \citep{vip}: Trains representations via an RL-based objective but has no language modality. We use pretrained DistilBERT~\citep{sanh2019distilbert} as the language encoder for VIP. $4)$ \textit{LIV} \citep{liv}: Extends VIP to vision-language representation learning using CLIP to align final images with instructions. $5)$ \textit{RT1} \citep{rt1}: An end-to-end LCBC method. We initialize RT1 with its original pretrained vision-language encoder for a fair comparison. We do not compare with Voltron~\citep{karamcheti2023language} as it is trained on small-scale Something-Something-v2~\citep{goyal2017something} data.

\begin{figure}[t]
    \centering
    \includegraphics[width=0.44\textwidth]{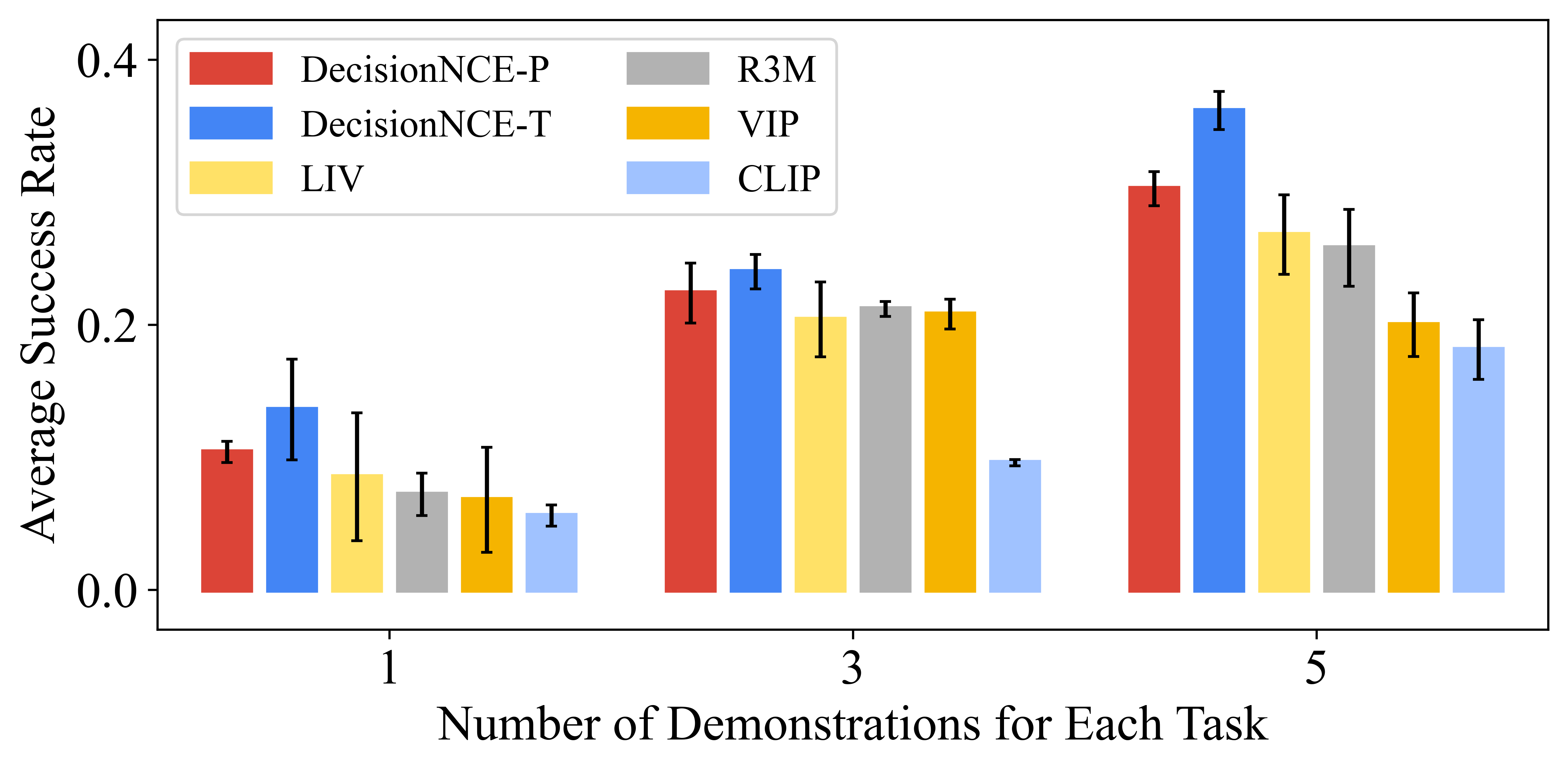}
     \vspace{-12pt}
    \caption{\small \textbf{Simulation LCBC results}. Max success rate averaged over 25 evaluation episodes and 3 seeds.}
    \label{fig:franka-imitation}
\end{figure}

\begin{figure*}[t]
    \centering
    \includegraphics[width=0.99\textwidth]{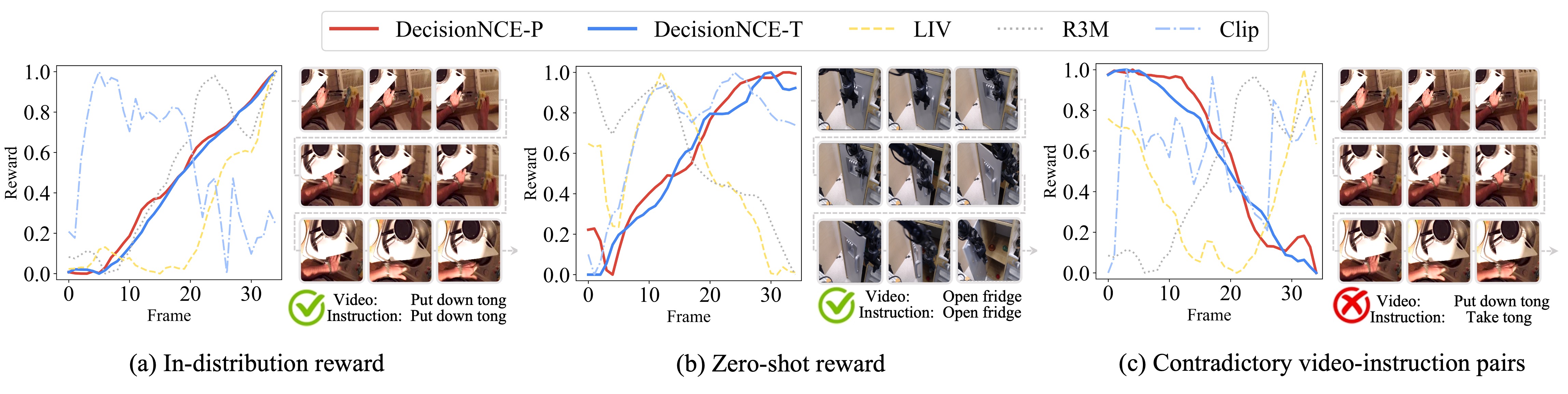}
    \vspace{-13pt}
    \caption{\small Visualization of the learned reward for diverse data. DecisionNCE provides more accurate rewards than baseline methods.}
    \label{fig:reward_curve}
\end{figure*}

\subsection{Language-conditioned Behavior Cloning Results}
\label{subsec:sim}
We freeze the pretrained vision-language encoders and use their output representations
as input to a 256-256 MLP to train LCBC policies, trained using a few domain-specific demonstrations with action annotations. Extensive results show that DecisionNCE can extract valuable information that facilitates efficient downstream policy learning.

\textbf{Results on real robots.} For real robots, we evaluate 9 distinct tasks.
For all tasks, the environment is highly stochastic with randomly initialized robots and objects, and contains lots of distractors with random locations, colors, and shapes. Ideally, the representations should be robust against these disturbances. Please see Appendix~\ref{appen:exp_details} for detailed setups.

Results are presented in Figure \ref{fig:realrobot}. DecisionNCE significantly outperforms baselines across almost all tasks. Among all baselines, LIV and R3M achieve the best results, benefiting from the alignment between images and languages as well as temporal consistency. CLIP, however, performs poorly as aligning languages with single images misses key task progression signals encoded in sequential data. Moreover, VIP cannot obtain satisfactory results as its pretrained vision and language embeddings are not aligned. RT1 also obtains subpar performances as the downstream data is too small to train a large model.

\textbf{Results on simulated robots.} We also evaluate on the FrankaKitchen~\citep{gupta2019relay} benchmark, following LIV~\citep{liv}. We train LCBC policies on 5 tasks in FrankaKitchen environment using 1/3/5 demonstrations for each task. Figure~\ref{fig:franka-imitation} shows that DecisionNCE achieves the highest success rate across diverse dataset quantities, demonstrating its effectiveness in extracting valuable information from out-of-domain data.

\subsection{Universal Reward Learning}
Note that the BT model is widely used in PBRL for reward learning, thus our learned representations can also be parameterized into universal reward signals (as in Section~\ref{subsec:reward_para}), offering direct guidance for accomplishing downstream tasks. We explore this ability of DecisionNCE by visualizing the normalized rewards calculated by image/language similarities on different videos. We also visualize rewards derived by other methods. Figure~\ref{fig:reward_curve}(a-b) demonstrates that DecisionNCE not only offers accurate reward guidance on in-distribution data, but also provides effective zero-shot rewards on out-of-distribution views and embodiments. Furthermore, Figure~\ref{fig:reward_curve}(c) shows DeicisionNCE is the only model that clearly identifies mismatched video/language pairs, revealing strong robustness against erroneous instructions compared to other methods. This robustness likely stems from DecisionNCE's training on randomly sampled segments (resists noisy transitions) and the use of contrastive learning against negative segments (enhances the ability to discern mismatched pairs).

\begin{table}[t]
\small
\setlength{\tabcolsep}{3.8pt}
\small 
    \centering
    \caption{\small Average success rate overall FrankaKitchen tasks using MPPI planning based on zero-shot implicit rewards.}
    \begin{tabular}{cccccccc}
    \toprule
     \multirow{2}{*}{Model} & \multirow{2}{*}{CLIP} & \multirow{2}{*}{R3M} & \multirow{2}{*}{LIV} & \multicolumn{2}{c}{DecisionNCE (Ours)} & ~\\
     \cmidrule(r){5-7}
     ~&~&~&~&-P&-T\\
     \midrule
     \thead{Average \\ Success Rate} & 0.016 & 0.036 & 0.054 &\colorbox{mine}{0.101} & 
     \colorbox{mine}{0.122}\\
     \bottomrule
    \end{tabular}
    \label{tab:mppi}
\end{table}
\subsection{Language-reward Planning}
To further assess the quality of implicit reward, we apply it in a model-based planning method, MPPI~\citep{williams2017model}, where future actions are optimized according to returns derived from the parameterized rewards, in a zero-shot manner. For baselines, we choose CLIP, LIV and R3M, and also use their customized rewards to run MPPI.
We conduct experiments on FrankaKitchen, adhering to experimental protocols from~\citet{vip} (See Appendix~\ref{frankakitchen} for details). Results in Table~\ref{tab:mppi} clearly show that DecisionNCE-P/T both significantly outperform other methods, indicating superior capability in capturing the similarity between the images and the language goals, thus offering more accurate guidance for generic language-directed tasks. Also, note that DecisionNCE-T performs better than DecisionNCE-P in most scenarios. We believe this is because DecisionNCE-T not only ensures correct relative transition directions, but also correct absolute positions, which makes it well-suited for reward/representation learning.


\section{Related Work}
\label{sec:related_work}
\textbf{Representation learning for decision making.} Pretrained representations are useful for downstream decision-making tasks~\citep{shah2021rrl, zhang2021learning, parisi2022unsurprising, cui2022can, laskin2020curl}. Yet, most prior works focus on narrow and expensive in-domain data~\citep{laskin2020reinforcement, yarats2021image, seo2022reinforcement, kumar2022pre, myers2023goal, bhateja2023robotic}. Our work explores a broader setting that leverages extensive cheap out-of-domain data like human demonstration videos~\citep{damen2018scaling, goyal2017something, grauman2022ego4d} to improve representation quality. While some works have studied similar settings, they mostly limit to uni-model representations that focus only on visual modality~\citep{mendonca2023structured}. For instance, \citet{xiao2022masked} and \citet{radosavovic2023real} noticed that Masked-Autoencoder~\citep{he2022masked, liu2023mixmae} for vision representation learning can improve decision making, and VIP~\citep{vip} performed Goal-Conditioned Reinforcement Learning (GCRL)~\citep{ma2022offline} with final image as goal specification to extract task-relevant features. 

\textbf{Language grounding for decision making.} 
Several studies directly apply existing multimodal representations, like CLIP~\citep{clip}, without tailoring them for decision making~\citep{radford2021learning, khandelwal2022simple, reid2022can}, potentially overlooking critical elements such as environmental dynamics and task progressions. For instance, LIV~\citep{liv} aligns language with static goal images using CLIP, missing these essential elements. By contrast, \textit{trajectory-level grounding}, which aligns language with image sequences, is more suited for decision making. Voltron~\citep{karamcheti2023language} proposed a joint vision-language objective, yet its temporal length is quite short, failing to capture global task progressions. Conversely, R3M~\citep{r3m} and LOReL~\citep{nair2022learning} employed a language reward head to extract global task progressions, but  miss some local transition details. HULC~\citep{mees2022matters} used CLIP to align language with image sequences, but is limited to short sequence length due to high computation costs.
Moreover, most previous works often employ additional time contrastive learning objective~\citep{sermanet2018time} for temporal consistency, leading to a complex trade-off between two stages: grounding and ensuring temporal consistency, balanced by a hard-to-tune hyper-parameter~\citep{r3m, liv}. In contrast, our DecisionNCE framework elegantly merges the two-stage objectives into a unified loss, bypassing all nuanced trade-offs, presenting a principled vision-language representation learning framework for decision-making. Please see Appendix~\ref{sec:related_works_appendix} for extended discussions.


\section{Conclusion}
\label{sec:conclusion}
In this paper, we show that by introducing several smart designs, i.e., {implicit preference annotations}, {random segment sampling}, and {reward reparameterization in embedding spaces}, the Bradley-Terry model in preference learning can be elegantly extended to vision-language representation learning. This adaptation naturally leads to an elegant and cohesive trajectory-level InfoNCE-style optimization objective that is specifically tailored for decision making.
The derived DecisionNCE framework effectively addresses the central challenges in {trajectory-level grounding} problem, offering a principled approach to universal representation and reward learning. 
Due to limited resources, we instantiate DecisionNCE based on the CLIP encoders. Future research can further scale up DecisionNCE to train large Vision-Language Models (VLMs).
More discussion on limitations and future direction can be found in Appendix~\ref{appen:limitation}. 

\section*{Impact Statement}
This paper presents work whose goal is to advance the field of Machine Learning. One potential ethical concern is that the abroad data used for representation pretraining may contain some potential privacy issues and biases. However, in this paper, we use the open-sourced EPIC-KITCHEN-100 human video dataset~\citep{damen2018scaling}, which has been well peer-reviewed, thus resolved this ethical concern. We also encourage researchers to filter out privacy-sensitive and biased data when they train DecisionNCE on their customized dataset.

\section*{Acknowledgement}
This work is supported by National Key Research and Development Program of China under Grant
(2022YFB2502904), and funding from Haomo.AI. The authors would like to thank Yeqi Feng and Wei Wang for discussions on the earlier draft of the work; The authors
would also like to thank the anonymous reviewers for their feedback on the manuscripts.

\nocite{langley00}

\bibliography{example_paper}
\bibliographystyle{icml2024}

\newpage
\appendix
\onecolumn
\section{Additional Discussions on Related Works}
\label{sec:related_works_appendix}
\subsection{In-depth Comparisons with Closely Related Studies}\label{sec:related_works_appendix_compare}
This section provides detailed comparisons with several closely related works, including R3M~\citep{r3m}, LIV~\citep{liv}, and Voltron~\citep{karamcheti2023language}.

\textbf{R3M}~\citep{r3m} designs two separate loss functions, one is time contrastive learning~\citep{sermanet2018time} to enforce temporal consistency and an alignment loss based on LOReL~\citep{nair2022learning} to achieve vision-language alignment:
\begin{equation}
\scriptsize
\begin{aligned}
    &\mathcal{L}_{\rm tcn}(\phi) = \frac{1}{B}\sum_{b=1}^B \left[-\log\frac{\exp(\mathcal{S}(\phi(o^b_i), \phi(o^b_j))}{\exp(\mathcal{S}(\phi(o^b_i), \phi(o^b_j))) + \exp(\mathcal{S}(\phi(o^b_i), \phi(o^b_k))) + \exp(\mathcal{S}(\phi(o^b_i), \phi(o^{\neq b}_j))} \right] \\
  &\mathcal{L}_{\rm alignment}(\phi, \theta) =\frac{1}{B}\sum_{b=1}^B \left[- 
         \log\frac{\exp(\mathcal{G}_{\theta}(\phi(o^{b}_{0}),\phi(o^{b}_{j});\psi(l^b)))}{\exp(\mathcal{G}_{\theta}(\phi(o^{b}_{0}),\phi(o^{b}_{j});\psi(l^b))) + \exp(\mathcal{G}_{\theta}(\phi(o^{b}_{0}),\phi(o^{b}_{i});\psi(l^b))) + \exp(\mathcal{G}_{\theta}(\phi(o^{\neq b}_{0}),\phi(o^{\neq b}_{j});\psi(l^{\neq b})))} \right] \\
    & \mathcal{L}_{\rm R3M}(\phi, \theta) = \lambda_{1} \mathcal{L}_{\rm tcn} + \lambda_{2} \mathcal{L}_{\rm alignment} + \lambda_{3}\|\phi(o_i)\|_1 + \lambda_{4}\|\phi(o_i)\|_2,
\end{aligned}
\label{equ:r3m}
\end{equation}

where $k>j>i$ are the frame ids of video $v^b$, $\mathcal{S}$ is negative L2 distance, $\mathcal{G}_{\theta}$ is a reward model and $\lambda_{1,2,3,4}$ are the weight parameters. For the alignment loss, the image $o^b_{0}$ is selected from the first 20\% of the video. R3M employs $o^b_{0}$ along with other randomly sampled images to align with language. 

See from Eq.~(\ref{equ:r3m}) that the choice of $o^b_{0}$ appears somewhat ad-hoc. This makes the task progression features extracted by R3M always limit to starting from the first image, which may overlook some intermediary transitions within a video. Moreover, R3M presents a two-stage training, separately considering the temporal consistency and language grounding, which inevitably introduces complex trade-offs among these two potential contradicting objectives. In addition, see from $\mathcal{L}_{\rm R3M}$ that R3M requires to balance a lot of hyperparameters of different loss, making it hard to tune. In contrast, DecisionNCE inherently merges the complex two-stage training processes into a simple and unified objective in Eq.~(\ref{equ:final_objective}) and Eq.~(\ref{equ:final_objective_T}), naturally marries language grounding and temporal consistency in a more principled and hyper-parameter-free manner, eliminating all hyper-parameters to balance these complex loss.

\textbf{LIV}~\citep{liv} trains temporally consistent vision representations using goal conditioned reinforcement learning based on VIP \citep{vip}, and utilizes CLIP~\citep{clip} to only align the static goal images with language:
\begin{equation}
\small
\begin{aligned}
    &\mathcal{L}_{\rm VIP}(\phi) = \frac{1-\gamma}{B}\sum_{b=1}^B \left[- \mathcal{S}(\phi(o^b_i), \phi(g^b)) \right] 
    + \log \frac{1}{B}\sum_{b=1}^B \exp \left[ \mathcal{S}(\phi(o^b_k), \phi(g^b)) + r - \gamma \mathcal{S}(\phi(o^b_{k+1}), \phi(g^b))\right]\\
  &\mathcal{L}_{\rm CLIP}(\phi, \psi) =\frac{1-\gamma}{B}\sum_{b=1}^B \left[ 
         -\log\frac{\exp (1-\gamma)\mathcal{S}(\phi(g^b),\psi(l^b))}{\frac{1}{B}\sum_{b'=1}^B \left[\exp (1-\gamma)\mathcal{S}(\phi(g^{b'}),\psi(l^{b'}))\right] }  \right] \\
    & \mathcal{L}_{\rm LIV}(\phi, \psi) = \lambda_1 \mathcal{L}_{\rm VIP} + \lambda_2 \mathcal{L}_{\rm CLIP},
\end{aligned}
\label{equ:liv}
\end{equation}
where $g^b$ is the last frame of video $v^b$. 
The employment of VIP loss aids in acquiring representations that adhere to temporal consistency. Nevertheless, aligning simply static goal images with language falls short of achieving trajectory-level grounding as single images cannot fully describe a dynamical behavior. In addition, see from Eq.~(\ref{equ:liv}) that LIV also faces a complex balance between temporal consistency and language grounding, but DecisionNCE solves this in an elegant and unified learning objective.

Moreover, note that LIV uses reinforcement learning to train representations, and thus must require correct reward signals. In LIV, the authors defined the rewards as 0 for goal frames and -1 for all other frames. However, it is widely known that hand-engineered reward design is challenging, and thus the human defined rewards are typically imperfect and may lead to unsatisfactory results~\citep{li2023mind}. In contrast, DecisionNCE is extended from the popular reward learning method, the Bradley-Terry model in Preference-based Reinforcement Learning or widely known Reinforcement Learning from Human Feedback (RLHF), to train representations, leading to fewer assumptions on rewards than LIV.


\textbf{Voltron}~\citep{karamcheti2023language} proposed two tasks, language conditioning and language generation, to jointly train images and language encoders. However, the selected image sequences are always short with a fixed length (the sequence length is fixed to only 2), leading to a lack of trajectory-level alignment. We do not compare with Voltron, as it is pre-trained solely on small-scale Something-Something-V2~\citep{goyal2017something} dataset.

\subsection{Language-Conditioned Decision Making}
To achieve generalist agent, another branch of works is end-to-end training language-conditioned vision-based agent~\citep{jang2022bc, reed2022generalist, rt1, rt2, shah2023mutex, shridhar2023perceiver}. These approaches, however, is quite data-hungry, requiring tremendous expensive in-domain data. Some works reuse the broad out-of-domain data to improve decision making via a hierarchical structure, such as vidio planning~\citep{black2023zero, du2023video, du2023learning, ajay2023compositional}, but does not show how representation affects decision making. DecisionNCE, however, effectively reuse out-of-domain data to extract decision-centric representations for efficient downstream policy learning, largely eliminating the significant demands for extensive task-specific data collection.

\subsection{Other Related Works}
We notice that trajectory-level grounding is closely related with action recognition and video prediction. Action recognition~\citep{shi2013sampling,wang2018temporal,korbar2019scsampler,zhi2021mgsampler,zhang2023closer} centers on detecting and categorizing movements of humans or objects in video sequences. Also, video prediction~\citep{oprea2020review,zakharov2022long} aims to forecast upcoming frames or events within a video. We can see that the core of these technologies also requires to extract local and global task progressions embedded in image sequences. Therefore, It would be an interesting topic to adopt the techniques from action recognition and video prediction to improve decision-centric vision-language representation learning.
\section{Proof of Eq.~(\ref{equ:strength})}
\label{appen:proof}
In this section, we briefly present the proof for Eq.~(\ref{equ:strength}). Here, we first review the conclusion in Eq.~(\ref{equ:strength}). In specific, the probability of image $o_t^+$ being selected as a goal is not isotropic but progressively increases as the video progresses. Mathematically, the probability for $o_t^+, 0<t<h$ being a goal image grows monotonically w.r.t its time stamp $t$:
$$P(o_t^+ {\rm \ is\ selected\ as\ goal})=\frac{1}{h}\sum_{i=1}^{i<t}\frac{1}{h-i}$$
\begin{proof}
\begin{equation}
    \begin{aligned}
        &P(o_t^+ {\rm \ is\ selected\ as\ goal})\\
        &\overset{\rm 1}{=}\sum_{i=1}^h P(o_i^+ {\rm \ is\ selected\ as\ start\ }\cap o_t^+{\rm \ is\ selected\ as\ goal\ when\ } o_i^+ {\rm \ is \ start})\\
        &\overset{\rm 2}{=}\sum_{i=1}^h P(o_i^+ {\rm \ is\ selected\ as\ start})\times P({\rm o_t^+ \ is\ selected\ as \ goal\ when\ } o_i^+{\rm \ is\ start})\\
        &\overset{\rm 3}{=}\frac{1}{h}\sum_{i=1}^{i<t}P({\rm o_t^+ \ is\ selected\ as \ goal\ when\ } o_i^+{\rm \ is\ start}) + \frac{1}{h}\sum_{i=1}^{t\leq i\leq h}P({\rm o_t^+ \ is\ selected\ as \ goal\ when\ } o_i^+{\rm \ is\ start})\\
        &\overset{\rm 4}{=}\frac{1}{h}\sum_{i=1}^{i<t}P({\rm o_t^+ \ is\ selected\ as \ goal\ when\ } o_i^+{\rm \ is\ start}) + 0\\
        &\overset{\rm 5}{=}\frac{1}{h}\sum_{i=1}^{i<t}\frac{1}{h-i},\\
    \end{aligned}
\end{equation}
where the 3rd equation holds since the start image $o_i$ is uniformly sampled from the whole video, and thus $P(o_i^+ {\rm \ is\ selected\ as\ start})=\frac{1}{h}$ where $h$ is the video length. The 4th equation holds because it is impossible to select $o_t^+$ as a goal image when $i>t$, as $o_t^+$ is selected from the following images after $o_i$ with $t>i$. The 5th equation holds because $o_t^+$ is uniformly selected from the following frames after $o_i$, and thus the remaining video length becomes $h-i$.
\end{proof}
\section{Theoretical Analysis (Informal)}
\label{appen:theory}

\textbf{What does DecisionNCE update do?} \quad For a mechanistic understanding of DecisionNCE, it is useful to analyze the gradient of its loss function $\mathcal{L}_{\text{DecisionNCE}}$. In the following, we present the gradient analysis of DecisionNCE-P (Definition~\ref{def:decision_P}) and DecisionNCE-T (Definition~\ref{def:decision_T}). We begin by introducing the main assumption used in our analysis.
\begin{assumption}[Differentiable and Lipschitz continuous function approximators]
    \textit{The vision encoder $\phi$ we trained is differentiable and $L$-Lipschitz continuous, i.e., $\|\nabla_x \phi(x) \| \leq L, \forall x \in \mathcal{O}.$}
\label{assum:lip}
\end{assumption}
Note that assumption \ref{assum:lip} is a mild assumption which is frequently utilized in plenty of works ~\citep{li2019towards, miyato2018spectral,virmaux2018lipschitz,liu2022learning}.

\textbf{DecisionNCE-P.}\quad According to the Definition \ref{def:decision_P} of the DecisionNCE-P: 
\begin{equation}
\small
\begin{aligned}
P_P\left[\sigma^+ \succ \sigma^-\right]&=\frac{\exp (\mathcal{S}(\phi(o_{n+m}^+);\psi(l^+)) - \mathcal{S}(\phi(o_n^+); \psi(l^+)))}{\exp(\mathcal{S}(\phi(o_{n+m}^+);\psi(l^+))-\mathcal{S}(\phi(o_n^+); \psi(l^+))) +\exp(\mathcal{S}(\phi(o_{n+m}^-); \psi(l^+))-\mathcal{S}(\phi(o_{n}^-);\psi(l^+)))} \\
&=\frac{1}{1+\exp(\mathcal{S}(\phi(o_{n+m}^-); \psi(l^+)) - \mathcal{S}(\phi(o_n^-);\psi(l^+)) - \mathcal{S}(\phi(o_{n+m}^+); \psi(l^+)) + \mathcal{S}(\phi(o_{n}^+); \psi(l^+)))} \\
\end{aligned}
\label{equ:extend_preference}
\end{equation}
The loss function of DecisionNCE-P is as follows:
\begin{equation}
\begin{aligned}
\small
    \mathcal{L}_{\text{DecisionNCE-P}} & (\phi,\psi) = - \log P_P\left[\sigma^+ \succ \sigma^-\right] \\
    &= \log \left[1+ \exp(\mathcal{S}(\phi(o_{n+m}^-); \psi(l^+)) - \mathcal{S}(\phi(o_{n}^-);\psi(l^+)) - \mathcal{S}(\phi(o_{n+m}^+); \psi(l^+)) + \mathcal{S}(\phi(o_{n}^+); \psi(l^+)))\right]
\end{aligned}
\end{equation}
The gradient with respect to the parameters $\theta$ of $\psi$ cab be written as:
\begin{equation}
\scriptsize
\begin{aligned}
& \nabla_{\theta} \mathcal{L}_{\text{DecisionNCE-P}}(\phi,\psi_{\theta}) \\
& =\left(1 - P_P\left[\sigma^+ \succ \sigma^-\right]\right) \left[ \left[\mathcal{S’}(\phi(o_{n+m}^-), \psi_{\theta}(l^+)) - \mathcal{S’}(\phi(o_{n}^-),\psi_{\theta}(l^+))\right]\nabla_{\theta}\psi_{\theta}(l^+) - \left[\mathcal{S’}(\phi(o_{n+m}^+), \psi_{\theta}(l^+)) - \mathcal{S’}(\phi(o_{n}^+), \psi_{\theta}(l^+))\right]\nabla_{\theta}\psi_{\theta}(l^+) \right] \\
& = \left(1 - P_P\left[\sigma^+ \succ \sigma^-\right]\right) \underbrace{\left[ \mathcal{S’}(\phi(o_{n+m}^-), \psi_{\theta}(l^+)) - \mathcal{S’}(\phi(o_{n}^-),\psi_{\theta}(l^+)) - \mathcal{S’}(\phi(o_{n+m}^+), \psi_{\theta}(l^+)) + \mathcal{S’}(\phi(o_{n}^+), \psi_{\theta}(l^+))\right]}_{g_P(o_n^-, o_{n+m}^-, o_n^+, o_{n+m}^+, l^+)}  \nabla_{\theta}\psi_{\theta}(l^+) \\
& = \left(1 - P_P\left[\sigma^+ \succ \sigma^-\right]\right) g_P(o_n^-, o_{n+m}^-, o_n^+, o_{n+m}^+, l^+) \nabla_{\theta}\psi_{\theta}(l^+)
\end{aligned}
\end{equation}
where we denote $\nabla_{b}\mathcal{S}(a,b)$ as $\mathcal{S}'(a,b)$. As we choose metric $\mathcal{S}$ as cosine similarity throughout the paper, $\mathcal{S}'(a,b) = \frac{a}{|a||b|} - \mathcal{S}\frac{b}{|b|^2}$. 
We perform a Taylor expansion on $ \mathcal{S'}(\phi(x),\psi_{\theta}(l^+))$ at $x = o_n$:
\begin{equation}
    \mathcal{S'}(\phi(x),\psi_{\theta}(l^+)) = \mathcal{S'}(\phi(o_n),\psi_{\theta}(l^+)) + \nabla_{\phi}\mathcal{S}'(\phi(x);\psi_{\theta}(l^+))\nabla_{x}\phi(x) |_{x=o_n}(x-o_n) + \mathcal{R}(x)
\end{equation}
where $R(x)$ is higher-order infinitesimal of $(x-o_n)$. Hence we have:
\begin{equation}
    \mathcal{S'}(\phi(o_n),\psi_{\theta}(l^+)) - \mathcal{S'}(\phi(o_{n+m}),\psi_{\theta}(l^+) =  -\nabla_{\phi}\mathcal{S}'(\phi(x);\psi_{\theta}(l^+))\nabla_{x}\phi(x) |_{x=o_n}(o_{n+m}-o_n) - \mathcal{R}(o_{n+m})
\end{equation}
And 
\begin{equation}
\begin{aligned}
    g_P(o_n^-, o_{n+m}^-, o_n^+, o_{n+m}^+, l^+) = &\nabla_{\phi}\mathcal{S}'(\phi(x);\psi_{\theta}(l^+))\nabla_{x}\phi(x) |_{x=o_n^-}(o_{n+m}^- - o_n^-) + \mathcal{R}(o_{n+m}^-) \\
    &- \nabla_{\phi}\mathcal{S}'(\phi(x);\psi_{\theta}(l^+))\nabla_{x}\phi(x) |_{x=o_n^+}(o_{n+m}^+ - o_n^+) - \mathcal{R}(o_{n+m}^+)
\end{aligned}
\end{equation}
where $\nabla_{a}S'(a,b) = \nabla_{a}\nabla_{b}S(a,b) = \nabla_{a} (\frac{a}{|a||b|} - \mathcal{S}\frac{b}{|b|^2})$ is bounded. According to assumption \ref{assum:lip}, $\nabla_x \phi(x)$ is bounded as well. When $m$ is set to a large value, there may be a substantial difference between $o_{n+m}$ and $o_n$, resulting in a situation where $g_P(o_n^-, o_{n+m}^-, o_n^+, o_{n+m}^+, l^+)$ contributes more to the loss gradient $\nabla_{\theta} \mathcal{L}_{\text{DecisionNCE-P}}(\phi,\psi_{\theta})$. In our work, $m$ is randomly set from 1 to the maximum segment length. In contrast, some other works (such as VIP-L in Equ.4 of \citep{liv}) resemble setting $m=1$, leading to a minor difference between $o_{n+m}$ and $o_n$, resulting in a slower change in the loss. This can be less effective in training language encoder compared to our method. 


\textbf{DecisionNCE-T.}\quad According to the definition \ref{def:decision_T} of the DecisionNCE-T: 

\begin{equation}
\small
\begin{aligned}
P_T\left[\sigma^+ \succ \sigma^-\right] & =\frac{\exp \left(\mathcal{S}(\phi(o_{n+m}^+)-\phi(o_{n}^+);\psi(l^+))\right)}{\exp \left(\mathcal{S}(\phi(o_{n+m}^+)-\phi(o_{n}^+);\psi(l^+))\right) + \exp \left(\mathcal{S}(\phi(o_{n+m}^-)-\phi(o_{n}^-);\psi(l^+))\right)} \\
& = \frac{1}{1+\exp\left( \mathcal{S}(\phi(o_{n+m}^-)-\phi(o_{n}^-);\psi(l^+)) - \mathcal{S}(\phi(o_{n+m}^+)-\phi(o_{n}^+);\psi(l^+)) \right)}
\end{aligned}
\end{equation}

The loss function of DecisionNCE-T is as follow:
\begin{equation}
\begin{aligned}
\small
    \mathcal{L}_{\text{DecisionNCE-T}} (\phi,\psi) &= - \log P_T\left[\sigma^+ \succ \sigma^-\right] \\
    &= \log \left[1+ \exp\left(\mathcal{S}(\phi(o_{n+m}^-)-\phi(o_{n}^-);\psi(l^+)) - \mathcal{S}(\phi(o_{n+m}^+)-\phi(o_{n}^+);\psi(l^+))\right)\right]
\end{aligned}
\end{equation}
The gradient with respect to the parameters $\theta$ of $\psi$ cab be written as:
\begin{equation}
\small
\begin{aligned}
\nabla_{\theta} \mathcal{L}_{\text{DecisionNCE-T}}(\phi,\psi_{\theta}) &=\left(1 - P_T\left[\sigma^+ \succ \sigma^-\right]\right) \underbrace{\left[ \mathcal{S}'(\phi(o_{n+m}^-)-\phi(o_{n}^-);\psi(l^+)) - \mathcal{S}'(\phi(o_{n+m}^+)-\phi(o_{n}^+);\psi(l^+))  \right]}_{g_T(o_n^-, o_{n+m}^-, o_n^+, o_{n+m}^+, l^+)} \nabla_{\theta}\psi_{\theta}(l^+) \\
& = \left(1 -  P_T\left[\sigma^+ \succ \sigma^-\right]\right) g_T(o_n^-, o_{n+m}^-, o_n^+, o_{n+m}^+, l^+) \nabla_{\theta}\psi_{\theta}(l^+)
\end{aligned}
\end{equation}
We perform Taylor expansion on $ \mathcal{S'}(\phi(x) - \phi(o_n),\psi_{\theta}(l^+))$ at $x=o_n$:
\begin{equation}
    \mathcal{S'}(\phi(x) - \phi(o_n),\psi_{\theta}(l^+)) = \mathcal{S'}(0,\psi_{\theta}(l^+)) + \nabla_{\phi}\mathcal{S}'(\phi(x) - \phi(o_n);\psi_{\theta}(l^+))\nabla_{x}\phi(x) |_{x=o_n}(x-o_n) + \mathcal{R}(x)
\end{equation}
where $\mathcal{R}(x)$ is higher-order infinitesimal of $(x-o_n)$. Hence $g_T$ can be written as:
\begin{equation}
\begin{aligned}
    g_T(o_n^-, o_{n+m}^-, o_n^+, o_{n+m}^+, l^+) &= \mathcal{S'}(0,\psi_{\theta}(l^+)) + \nabla_{\phi}\mathcal{S}'(\phi(x) - \phi(o_n^-);\psi_{\theta}(l^+))\nabla_{x}\phi(x) |_{x=o_n^-}(o_{n+m}^- -o_n^-) + \mathcal{R}(o_{n+m}^-) \\
    & \quad \, -\mathcal{S'}(0,\psi_{\theta}(l^+)) - \nabla_{\phi}\mathcal{S}'(\phi(x) - \phi(o_n^+);\psi_{\theta}(l^+))\nabla_{x}\phi(x) |_{x=o_n^+}(o_{n+m}^+ -o_n^+) - \mathcal{R}(o_{n+m}^+) \\
    &= \nabla_{\phi}\mathcal{S}'(\phi(x) - \phi(o_n^-);\psi_{\theta}(l^+))\nabla_{x}\phi(x) |_{x=o_n^-}(o_{n+m}^- -o_n^-) + \mathcal{R}(o_{n+m}^-) \\
    & \quad \, -\nabla_{\phi}\mathcal{S}'(\phi(x) - \phi(o_n^+);\psi_{\theta}(l^+))\nabla_{x}\phi(x) |_{x=o_n^+}(o_{n+m}^+ -o_n^+) - \mathcal{R}(o_{n+m}^+) \\
\end{aligned}
\end{equation}
Similarly, in our method, $m$ is randomly set from 1 to the maximum segment length, implying a faster descent in loss compared to setting $m=1$ alone.



\section{Additional Results}
\label{appen:results}
In this section, we provide more cases to demonstrate that our DecisionNCE-P/T can learn accurate and generalizable reward, and conduct a deeper analysis.

\subsection{In-distribution reward}
\label{appen:indis-re}
\begin{figure}[htbp]
    \centering
    \subfigure[]{
        \includegraphics[width=3in]{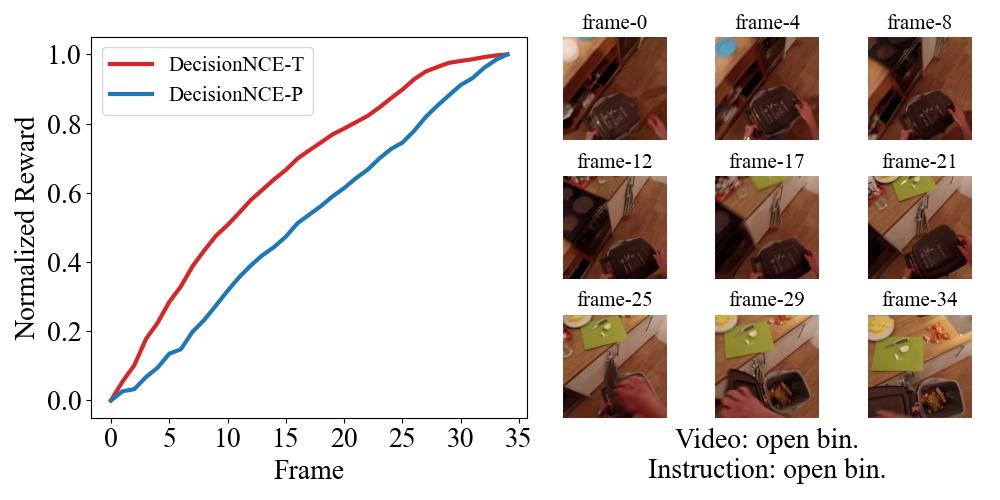}
    }
    \subfigure[]{
	\includegraphics[width=3in]{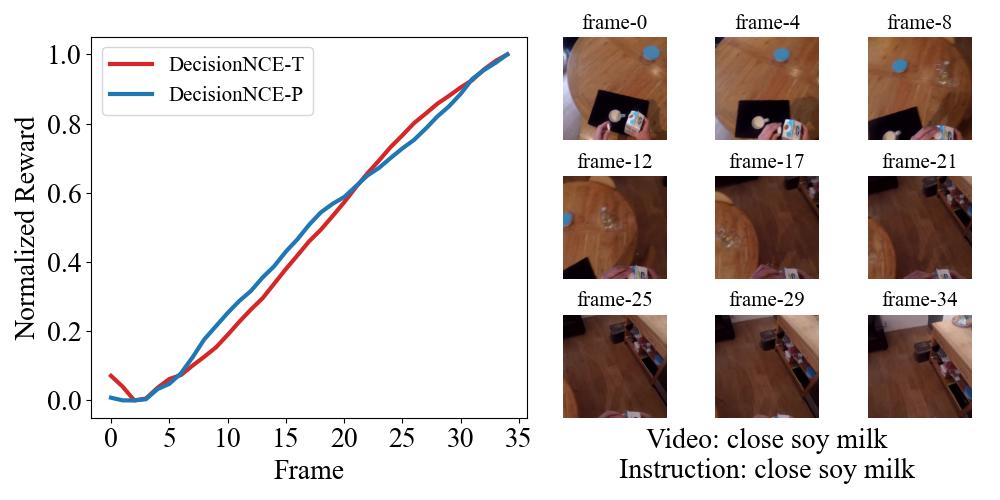}
    }
    \quad    
    \subfigure[]{
    	\includegraphics[width=3in]{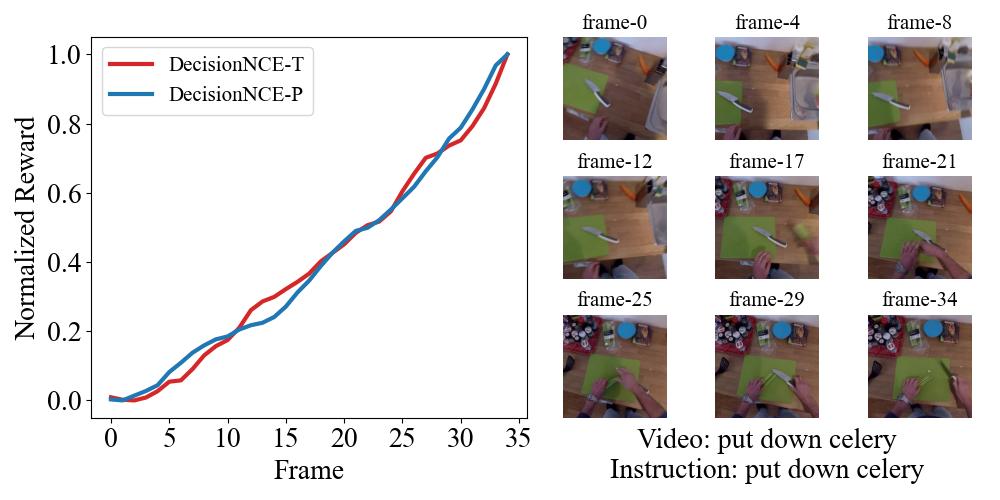}
    }
    \subfigure[]{
	\includegraphics[width=3in]{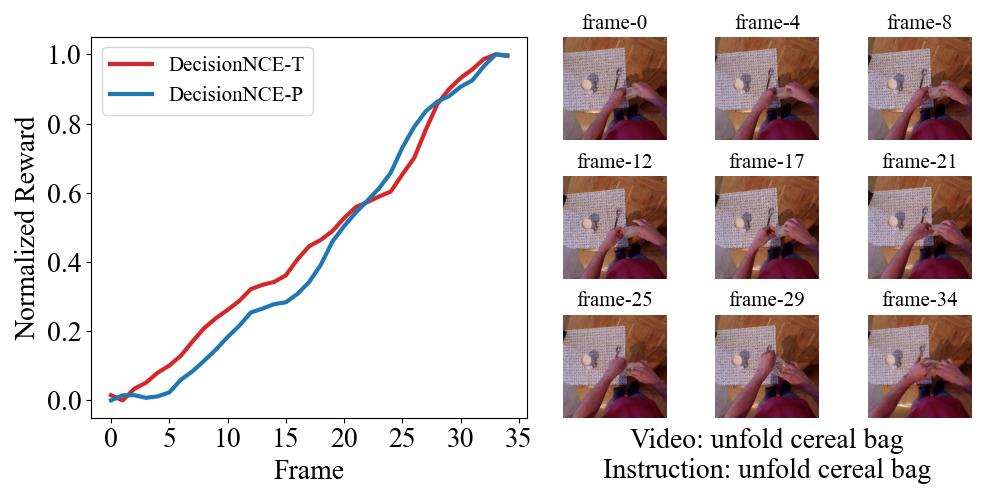}
    }
    \quad    
    \subfigure[]{
    	\includegraphics[width=3in]{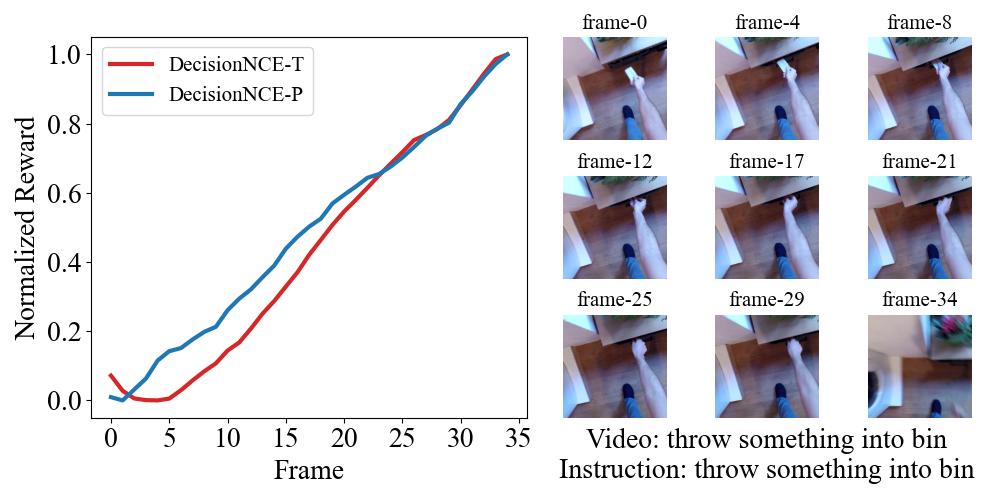}
    }
    \subfigure[]{
	\includegraphics[width=3in]{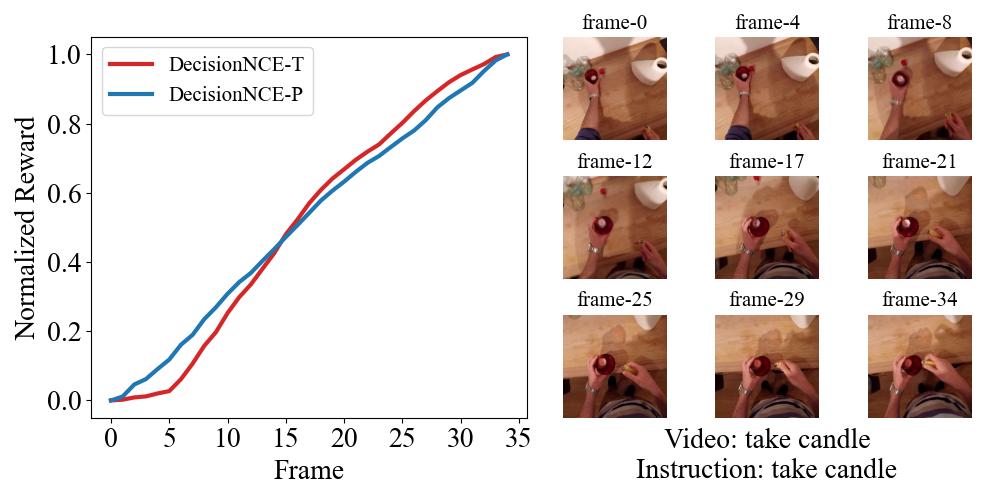}
    }
    \quad    
    \subfigure[]{
    	\includegraphics[width=3in]{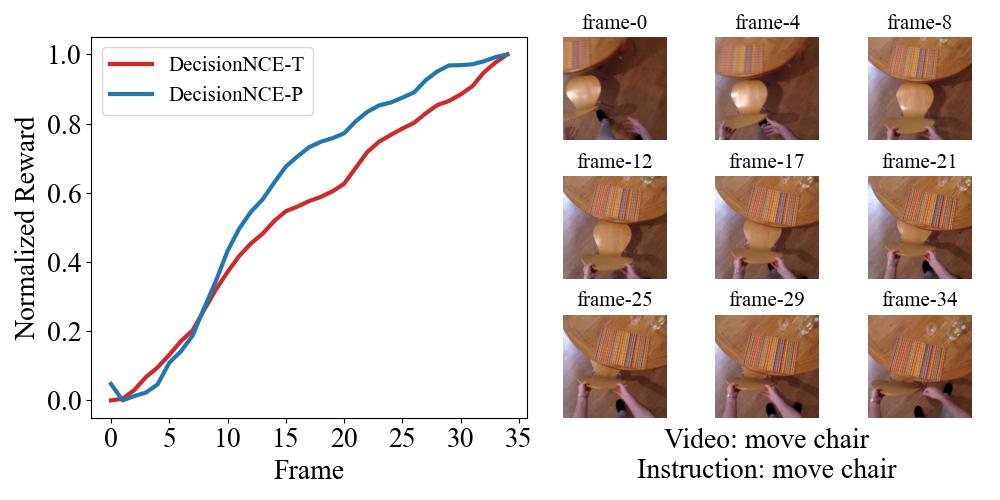}
    }
    \subfigure[]{
	\includegraphics[width=3in]{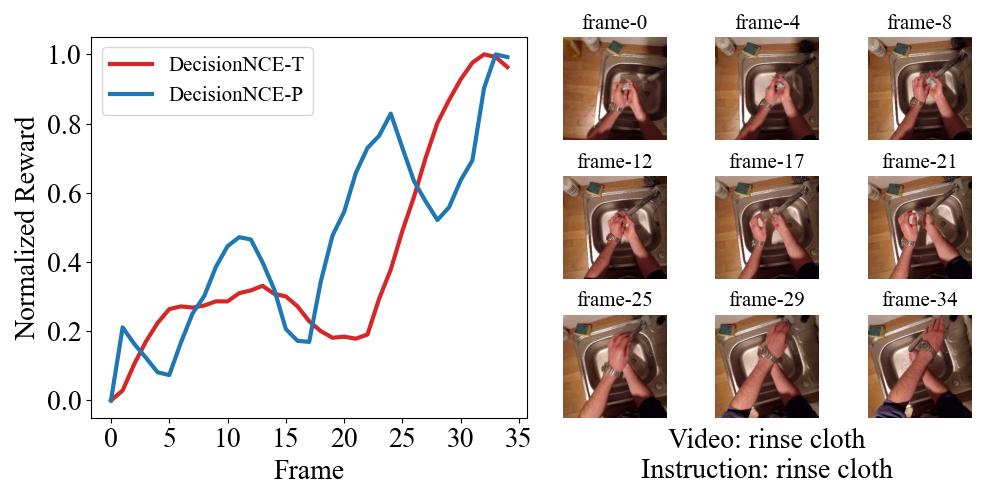}
    }
    \caption{In-distribution Reward}
    \label{fig.in-dis-reward}
\end{figure}

We randomly selected segments from EPIC-KITCHEN-100 dataset~\citep{damen2018scaling} with their corresponding text annotations to visualize the rewards curves learned through DecisionNCE-P/T. As shown in Figure~\ref{fig.in-dis-reward}, DecisionNCE accurately assigns rewards to different video frames across various action types, scenes, and target objects, resulting in a smooth and trend-correct reward curve. Next, we will delve deeper into discussing some intriguing and insightful characteristics observed in specific reward curves.

\textbf{Slope of curves:} 
The slope of the reward curves provides an insight into the task's progression.  For example, curves $(c)$ and $(g)$ demonstrate varying growth rates. This variation is due to the differing amplitudes of actions in the videos over time, as can be seen in the accompanying images.

\textbf{Robustness in Complex Tasks:}
In the context of kitchen tasks, which may involve complex actions with subgoals such as \texttt{drying hands}, \texttt{rinsing a cloth}, or \texttt{cutting something}, the robustness of DecisionNCE is notable. As illustrated by curve $(h)$, the DecisionNCE curve fluctuates, reflecting the presence of subgoals. This variation is a reasonable response to the complexity of the tasks.

\subsection{Zero-shot reward}

\begin{figure}[htbp]
    \centering
    \subfigure[]{
        \includegraphics[width=3in]{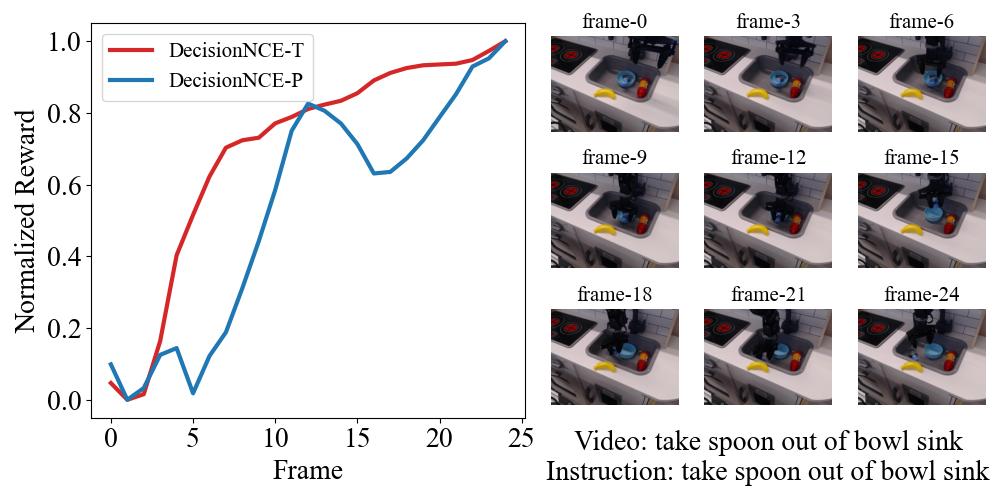}
    }
    \subfigure[]{
	\includegraphics[width=3in]{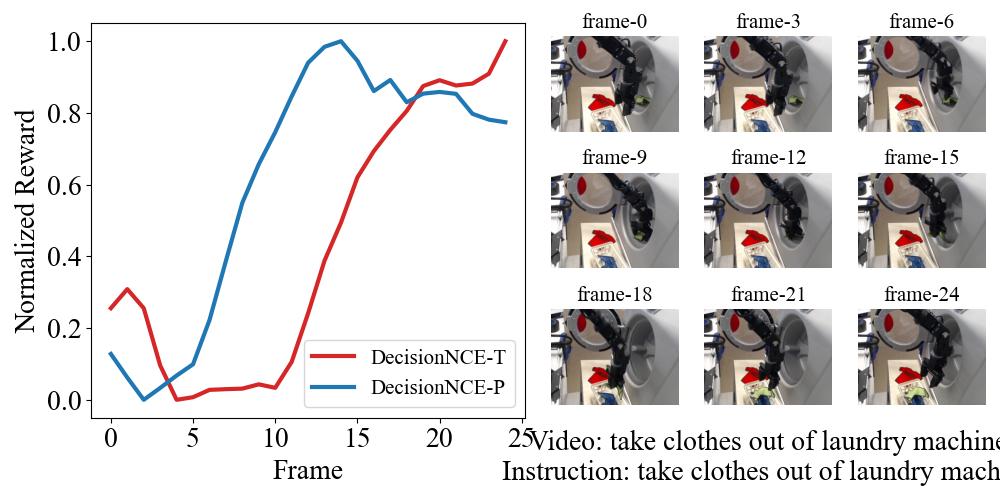}
    }
    \quad    
    \subfigure[]{
    	\includegraphics[width=3in]{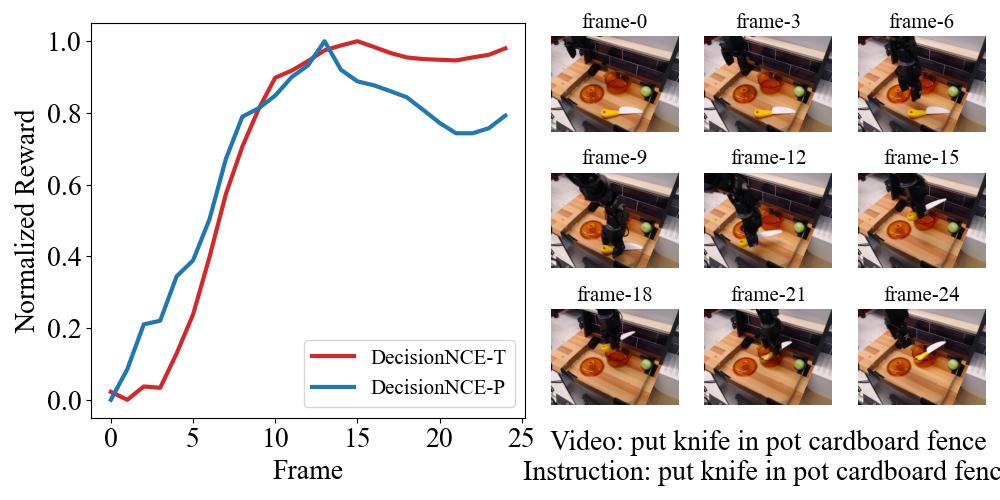}
    }
    \subfigure[]{
	\includegraphics[width=3in]{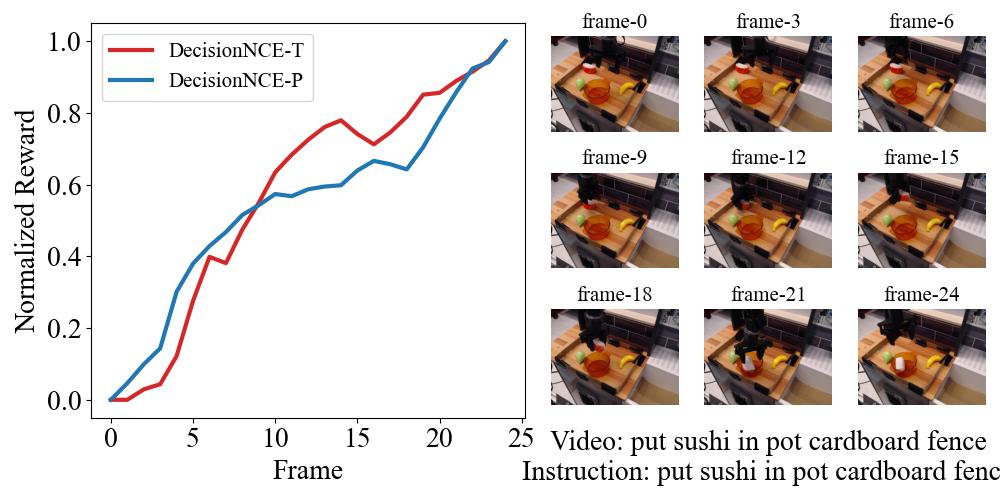}
    }
    \quad    
    \subfigure[]{
    	\includegraphics[width=3in]{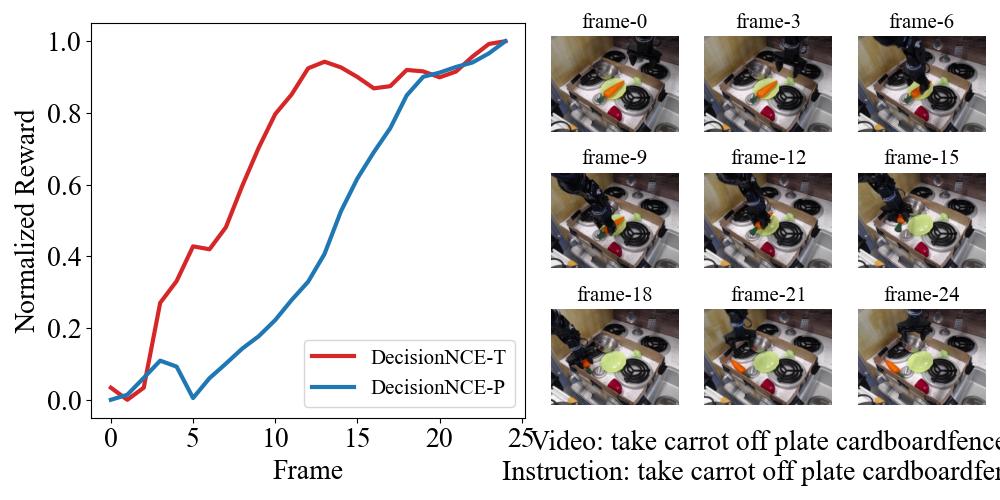}
    }
    \subfigure[]{
	\includegraphics[width=3in]{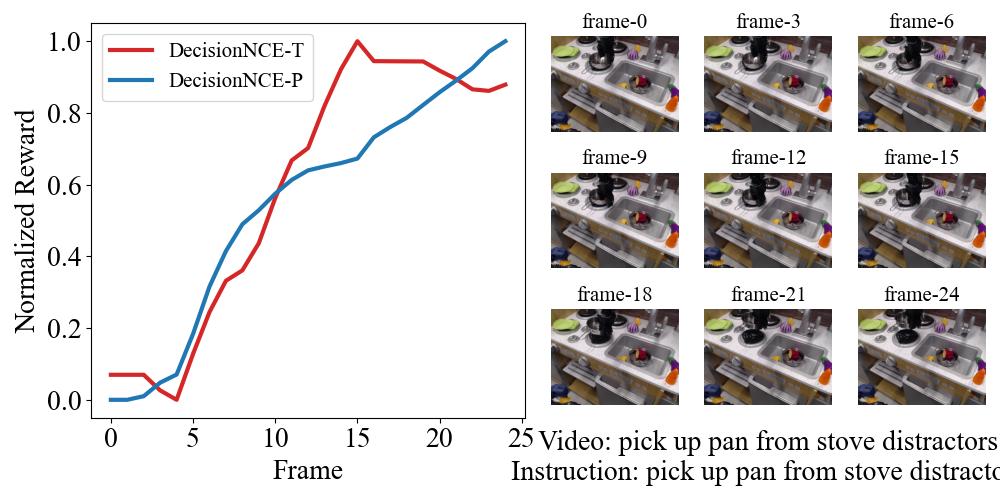}
    }
    \quad    
    \subfigure[]{
    	\includegraphics[width=3in]{figure/appendix/Bridgedata/put_knife_in_pot_cardboard_fence.jpg}
    }
    \subfigure[]{
	\includegraphics[width=3in]{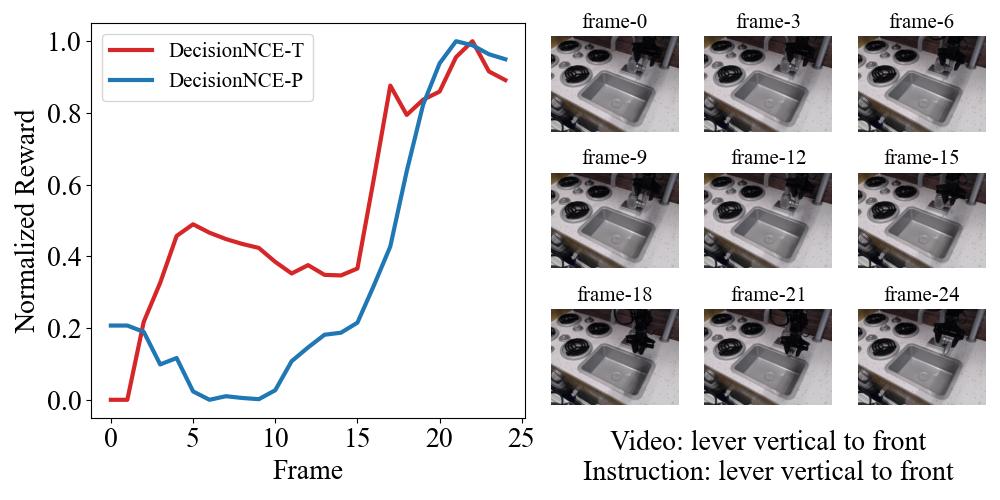}
    }
    \caption{Zero-shot Reward}
    \label{fig.zero-shot-reward}
\end{figure}

To assess the generalizability of the reward learned through DecisionNCE, we also visualize the reward curves on Bridgedata-V2 ~\cite{walke2023bridgedata},  a open-source robotics manipulation dataset. The results are presented in Figure~\ref{fig.zero-shot-reward}. DecisionNCE successfully captures the correct trend, demonstrating promising generalizability on out-of-domain data. It is important to note the significant domain gap between EPIC-KITCHEN-100 and Bridgedata-V2 which encompasses differences in embodiments, scenes, and view points.

\subsubsection{Failure cases analysis}

\begin{figure}[htbp]
    \centering
    \subfigure[]{
        \includegraphics[width=3in]{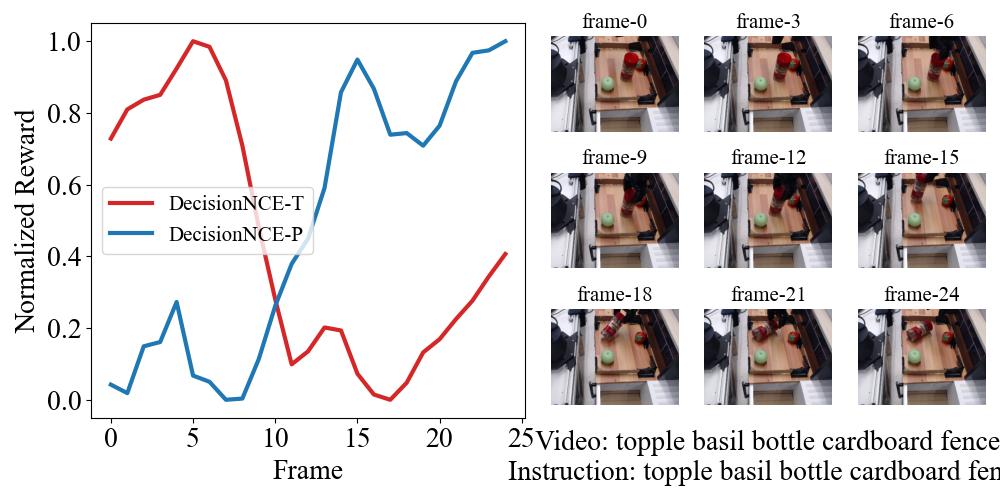}
    }
    \subfigure[]{
	\includegraphics[width=3in]{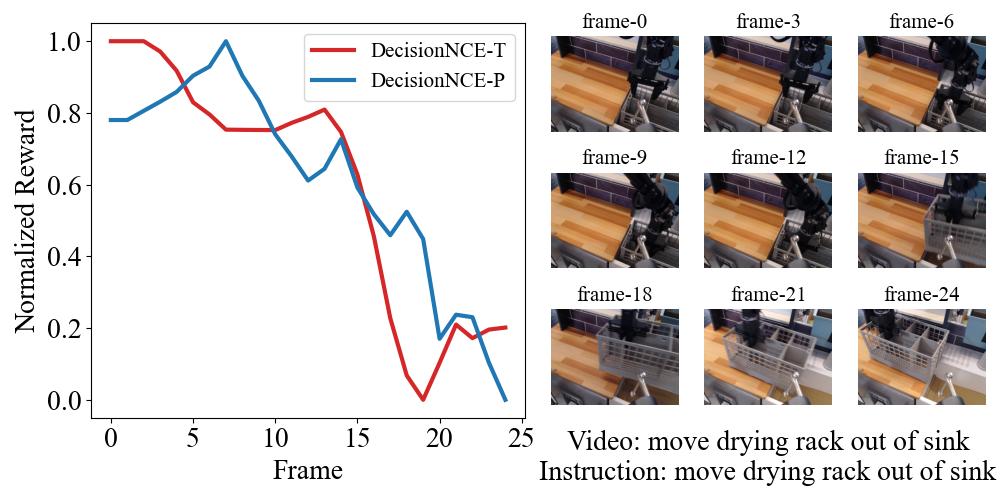}
    }

    \caption{Zero-shot failure cases}
    \label{fig.zero-shot-fail-reward}
\end{figure}

Indeed, while we observe numerous successful instances, there are also several results that deviate from our expectations. Upon a thorough examination of these unsuccessful cases, we discern two main factors. To provide a clearer understanding of these issues, we select two representative cases as shown in Figure~\ref{fig.zero-shot-fail-reward} for in-depth analyses.

\textbf{Out-of-domain skills}. 
Understanding actions at the trajectory level is notably more challenging to transfer than image perception skills. As a result, DecisionNCE demonstrates impressive performance with in-domain skills like \texttt{pick-up}, \texttt{put-down}, \texttt{open}, and \texttt{take-out}, maintaining its effectiveness even when there are changes in the target object or action sequence. However, it struggles when confronted with instructions describing a novel skill not previously encountered in the Epic-Kitchen dataset. For instance, in the case of \texttt{Topple} as shown in $(a)$, DecisionNCE is unable to accurately track the correct task progression.

\textbf{Dataset-specific task bias}.
The lack of precision in text annotations within the EPIC-KITCHEN-100 dataset results in a dataset-specific task bias, as learned by DecisionNCE. A prime example of this can be seen in $(b)$, where the task \texttt{move drying rack out of sink} exhibits an incorrect reward trend according to DecisionNCE's learning. This issue arises crucially due to a distinctive pattern in EPIC-KITCHEN-100. In EPIC-KITCHEN-100 dataset, most \texttt{move} actions result in the target object moving away from the center of the field of view. However, in this Bridgedata-V2 example, the scenario is reversed, which contributes to this erroneous reward curve.

In conclusion, the observed failures are primarily attributed to the limitations in the quality and scope of the dataset. For future advancements, it is imperative to enhance these aspects by incorporating a broader, more diverse, and higher-quality range of pre-training datasets. Please see Appendix~\ref{appen:limitation} for more discussions on limitations and solutions.

\subsection{Reward for mismatched video-instruction pair}

As metioned in Section~\ref{sec:experiments}, DecisionNCE effectively identifies mismatched video-language pairs. In this section, we delve deeper into the rewards assigned by DecisionNCE to these mismatched pairs, offering a detailed discussion of our findings.

\subsubsection{Completely irrelevant video-instructions pairs}

\begin{figure}[htbp]
    \centering
    \subfigure[]{
        \includegraphics[width=3in]{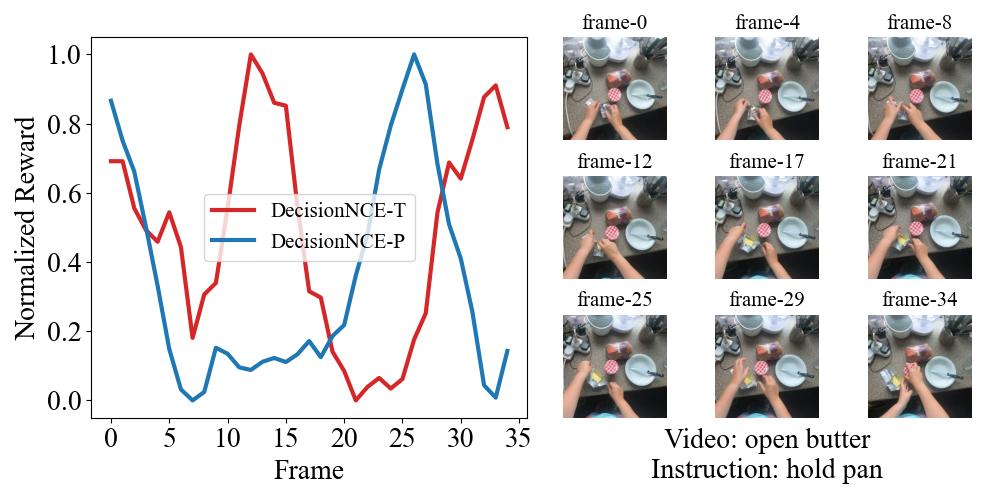}
    }
    \subfigure[]{
	\includegraphics[width=3in]{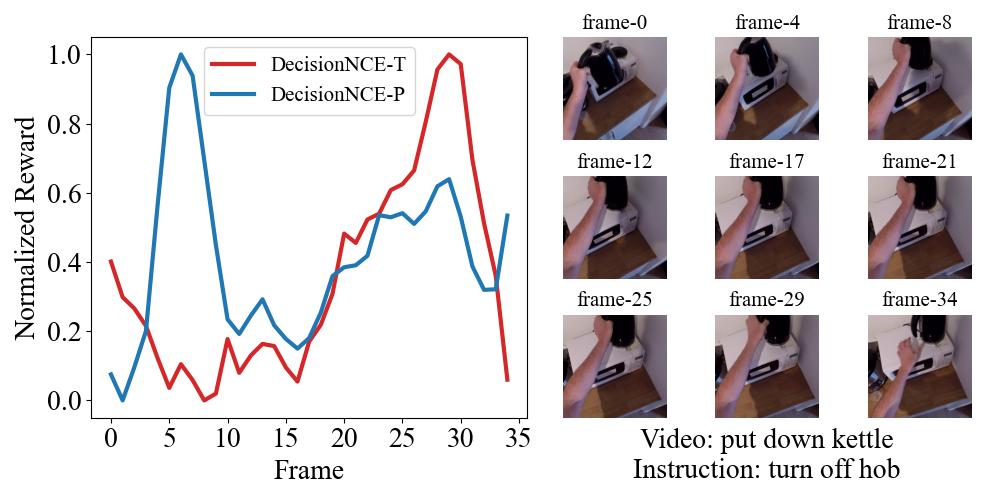}
    }
    \quad    
    \subfigure[]{
    	\includegraphics[width=3in]{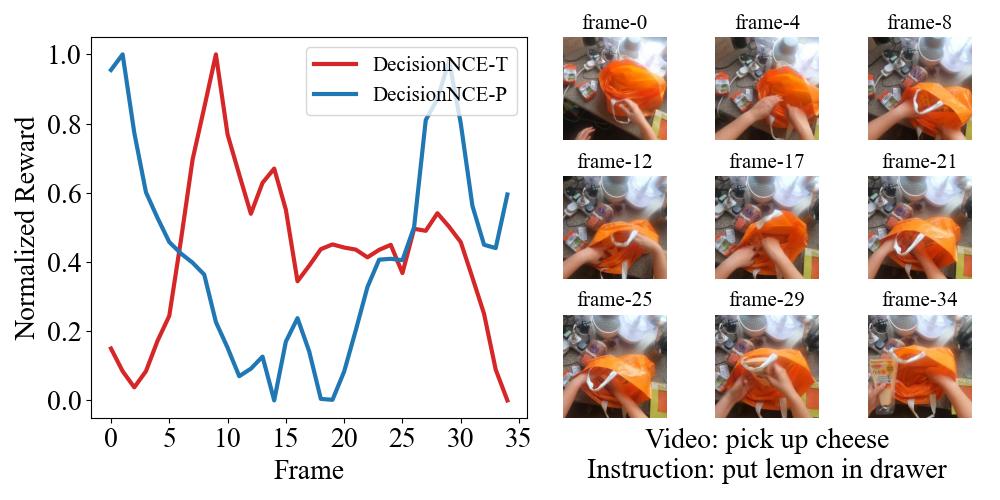}
    }
    \subfigure[]{
	\includegraphics[width=3in]{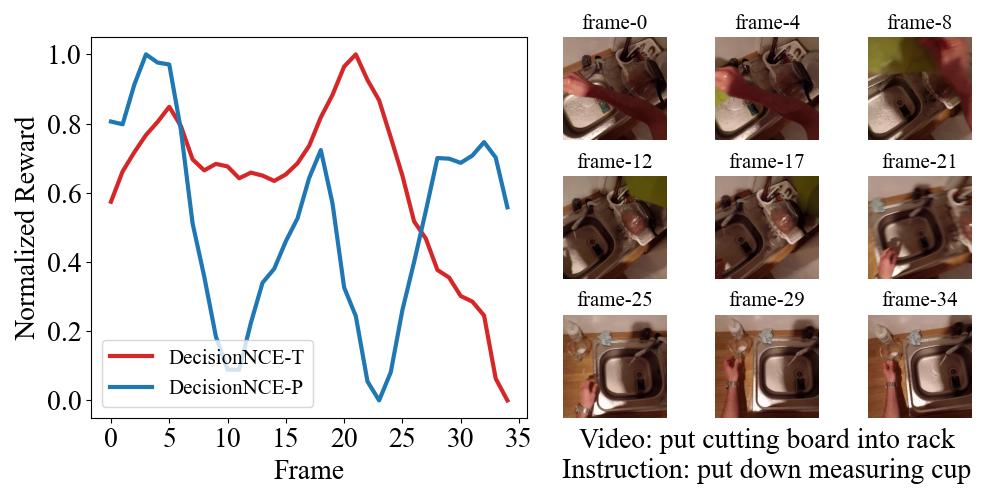}
    }
    \quad    
    \subfigure[]{
    	\includegraphics[width=3in]{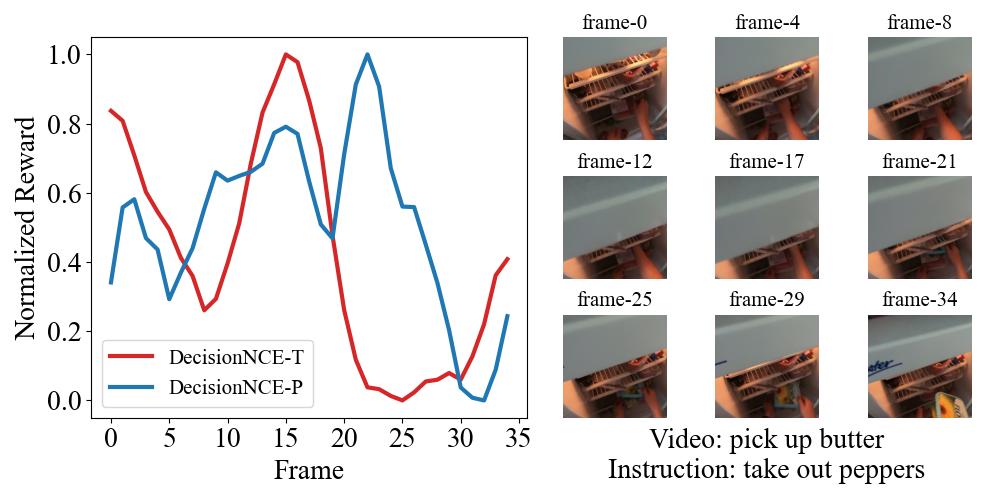}
    }
    \subfigure[]{
	\includegraphics[width=3in]{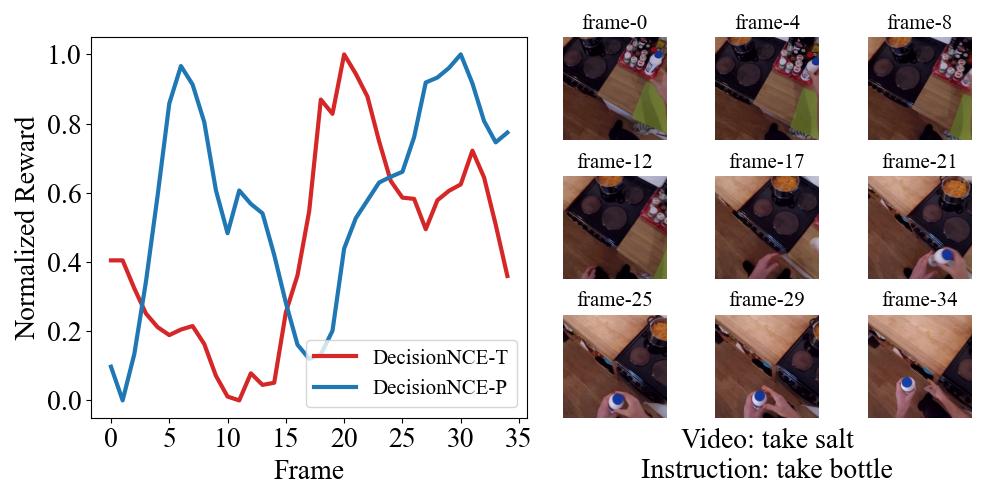}
    }
    \caption{Irrelevant video-instruction pair reward}
    \label{fig.unrelated-reward}
\end{figure}

For most cases, altering either the target object or the action in the instruction renders it incongruent with the task progression shown in the video. Under these \textbf{Completely Irrelevant Video-Instructions Pairs}, the resulting reward curve tends to lack a clear trend and appears chaotic, as illustrated in the Figure~\ref{fig.unrelated-reward}. Significantly, DecisionNCE does not erroneously interpret these unmatched instructions as correct, even when there is similarity in actions or objects. This demonstrates the model's strong robustness in differentiating between matched and mismatched pairs.

Certainly, mismatched video-instruction pairs are not always entirely irrelevant. Next, we will explore two intriguing cases to further illustrate the generalizability of DecisionNCE.

\subsubsection{Contradictory video-instructions pairs}
\begin{figure}[htbp]
    \centering
    \subfigure[]{
        \includegraphics[width=3in]{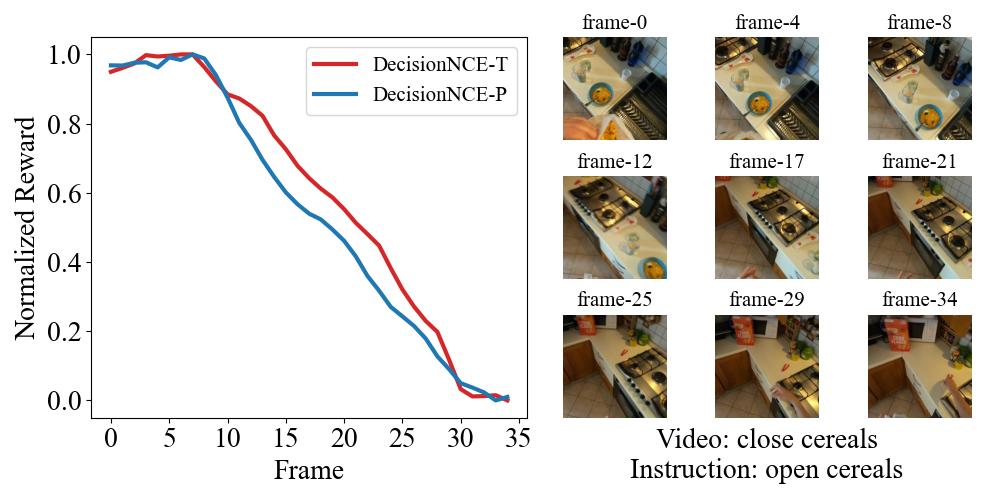}
    }
    \subfigure[]{
	\includegraphics[width=3in]{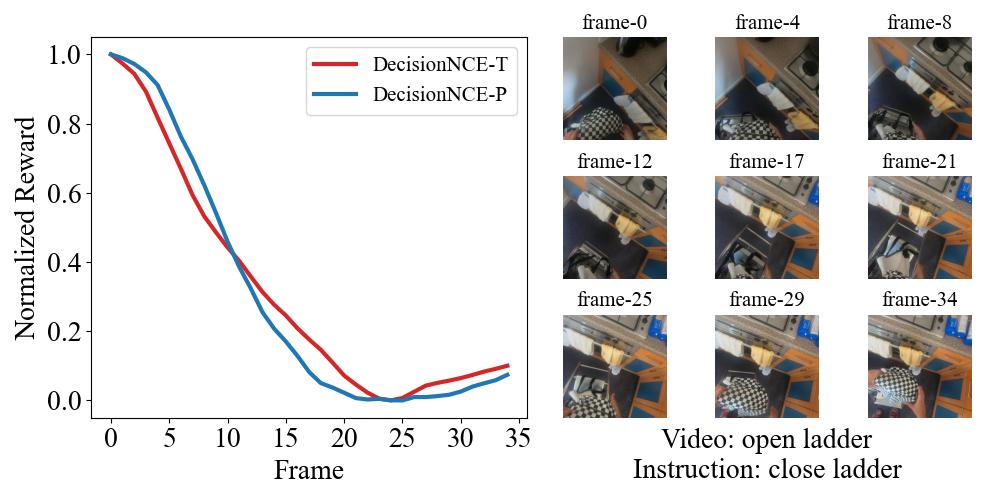}
    }
    \quad    
    \subfigure[]{
    	\includegraphics[width=3in]{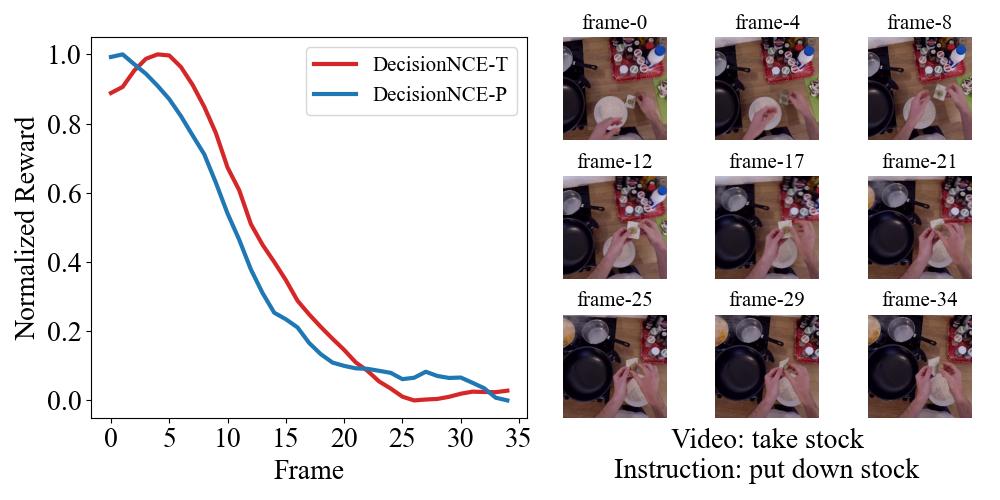}
    }
    \subfigure[]{
	\includegraphics[width=3in]{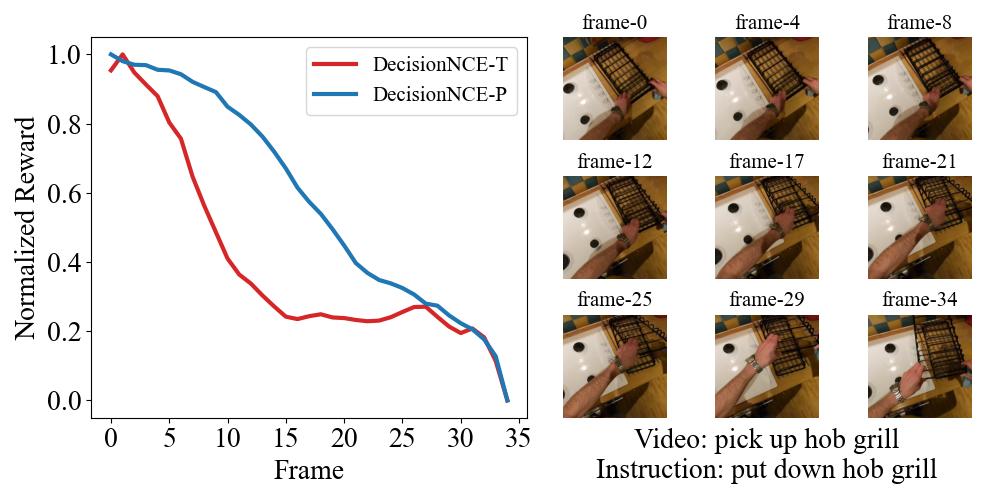}
    }
    \quad    
    \subfigure[]{
    	\includegraphics[width=3in]{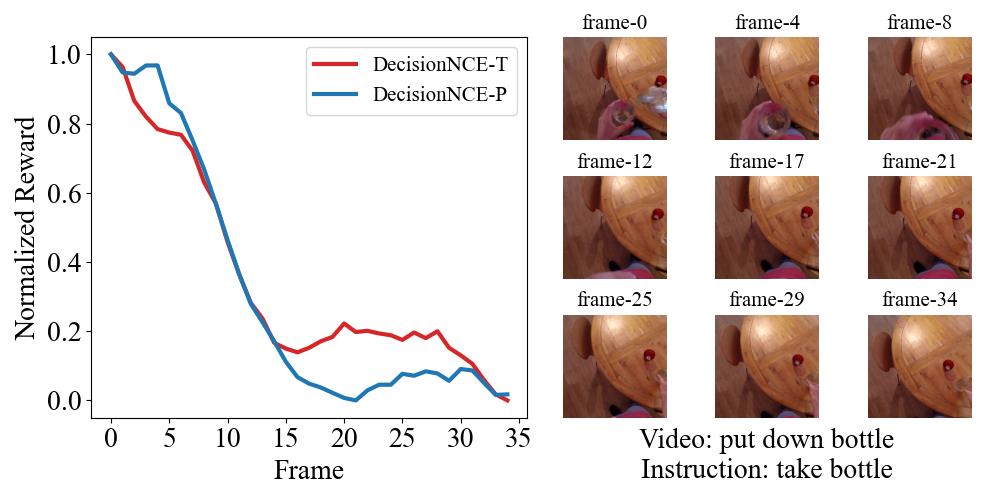}
    }
    \subfigure[]{
	\includegraphics[width=3in]{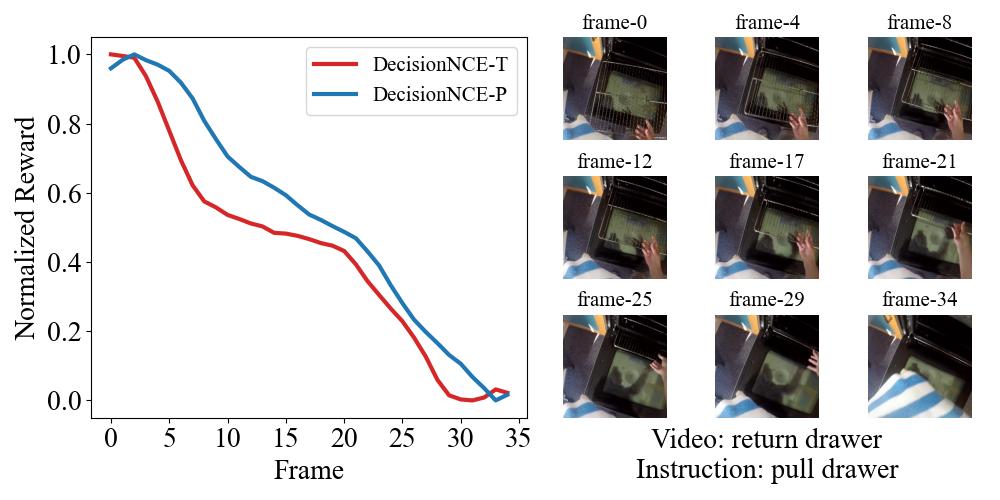}
    }
    \caption{Contradictory video-instruction pair reward}
    \label{fig.contradictory-reward}
\end{figure}

Given that tasks in EPIC-KITCHEN-100 are predominantly single-step, many have corresponding mirror tasks, such as \texttt{Open} and \texttt{Close}, \texttt{Pick-up} and \texttt{Put-down}, \texttt{Pull} and \texttt{Return}. With these mirrored instructions, the task progression in the video is effectively the opposite. We have visualized the reward curve for these \textbf{Contradictory Video-Instruction Pairs} in a similar manner as before. As depicted in Figure~\ref{fig.contradictory-reward}, the reward curve aligns with our expectations: as the task progresses, the reward value tends to decrease.

\begin{figure}[htbp]
    \centering
    \subfigure[]{
        \includegraphics[width=3in]{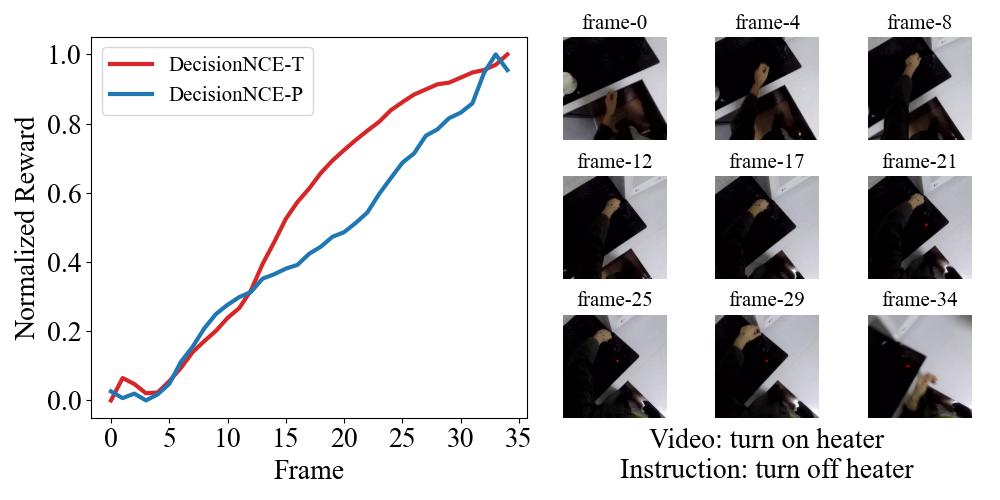}
    }
    \subfigure[]{
	\includegraphics[width=3in]{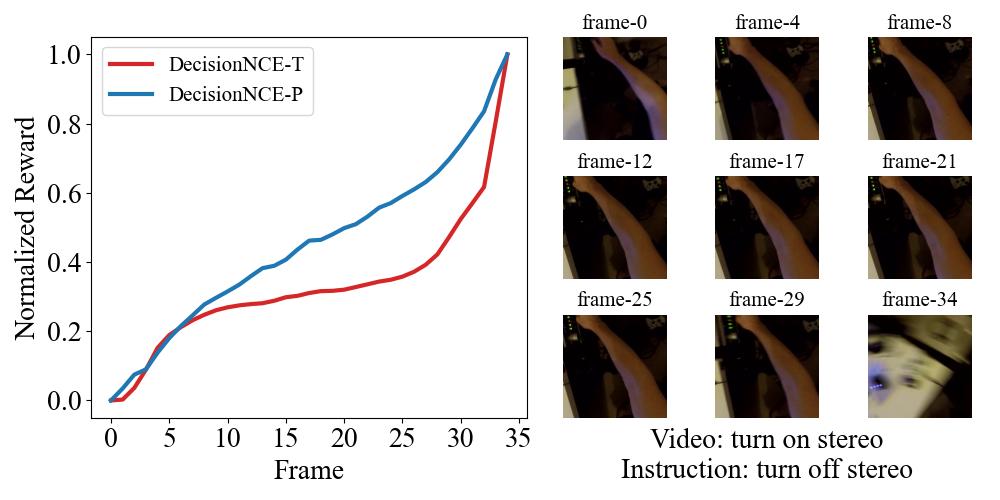}
    }

    \caption{Failure cases of contradictory pairs}
    \label{fig.contradictory-fail-reward}
\end{figure}

\textbf{Failure cases: }
We also discovered that certain mirror tasks pose significant challenges in differentiation from their original counterparts. As shown in Figure~\ref{fig.contradictory-fail-reward}, tasks such as \texttt{Turn on} and \texttt{Turn off} an appliance, like a heater $(a)$ or stereo $(b)$, often involve almost identical movement trajectories. To accurately distinguish between these tasks, the model needs to discern the initial and final states of the target appliance, a task that becomes exceedingly difficult without supplemental annotations. In this instance, DecisionNCE fails to differentiate between these two tasks.

\subsubsection{Indirectly relevant video-instructions pairs}

\begin{figure}[htbp]
    \centering
    \subfigure[]{
        \includegraphics[width=3in]{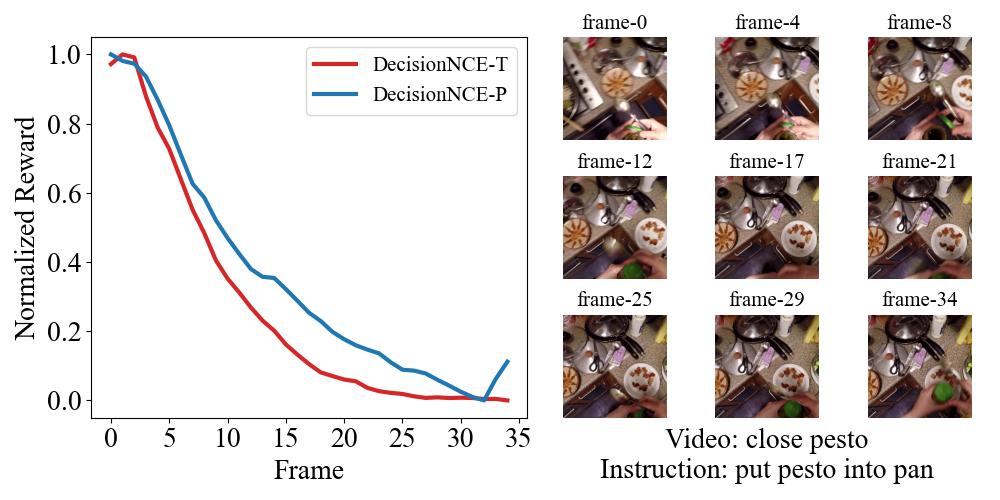}
    }
    \subfigure[]{
	\includegraphics[width=3in]{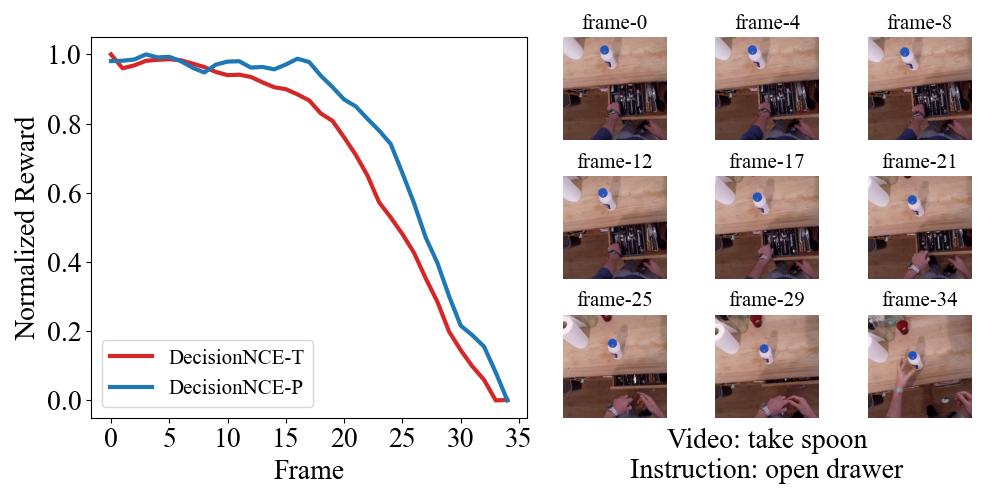}
    }

    \quad
    \subfigure[]{
        \includegraphics[width=3in]{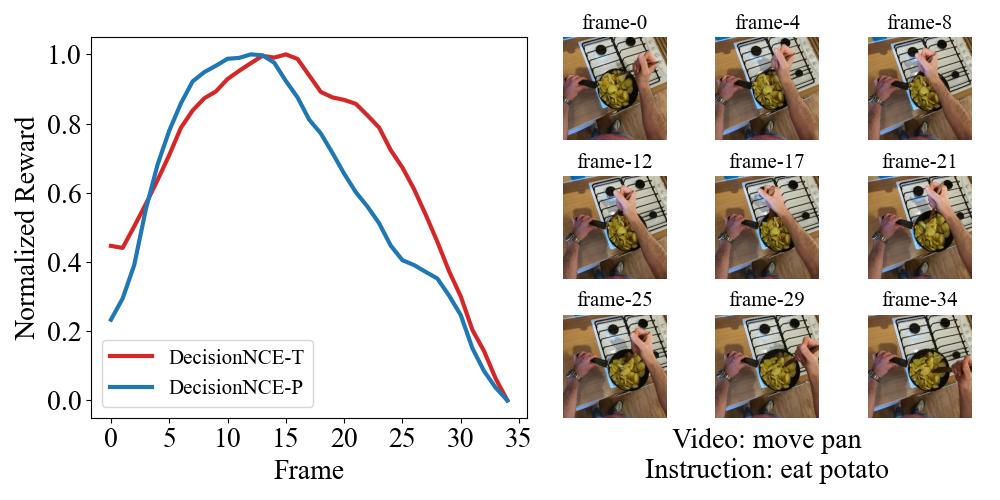}
    }
    \subfigure[]{
	\includegraphics[width=3in]{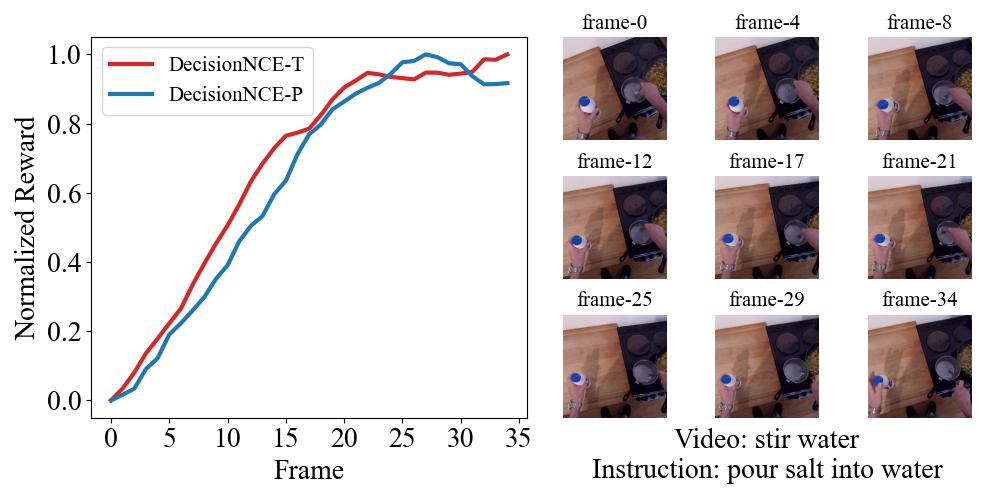}
    }
    \caption{Indirectly relevant video-instructions pairs reward }
    \label{fig.indirect-related-reward}
\end{figure}

Usually, the completion of a task typically involves accomplishing several sub-tasks or certain task subgoals. The Epic-Kitchen dataset, however, does not provide annotations with such granularity. For these sub-task instructions, they are indirectly related to the video. We have illustrated the reward curve for these \textbf{Indirectly Relevant Video-Instruction Pairs} in Figure~\ref{fig.indirect-related-reward}. The results show that DecisionNCE effectively recognizes these sub-tasks, even in the absence of explicit human annotations. For instance, in the case of video (b), the annotation is \texttt{take spoon}, yet the task \texttt{close drawer} is concurrently completed. So when \texttt{open drawer} is used as the instruction, DecisionNCE appropriately assigns a reward curve with an opposing trend.

\section{Experimental setups}
\label{appen:exp_details}
In this section, we provide detailed experimental setups. 

\subsection{Vision-Language representation training}
\label{sub_app:pretrain}
Following LIV~\cite{liv}, we use a modified ResNet-50~\cite{he2016resnet} from CLIP~\cite{clip} for the vision encoder and a CLIP transformer for the language encoder. We initialize our DecisionNCE-P/T with CLIP model and train them on EPIC-KITCHEN-100~\citep{damen2018scaling} under the same training setting. The training hyperparameters used during the pre-training are listed in Table~\ref{tab:pretrain-setting}. It is worth noting that our training objective in Algorithm~\ref{algo:decisionNCE} compares all mismatched pairs in mini-batch samples, which enjoys much higher training efficiency and thus only 20K training steps can obtain well-trained representations. Therefore, our training only take about 9 hours on four A100 GPUs, showing higher training efficiency compared to previous work.

\begin{table}[h]
\caption{Hyper-parameters for pretraining .}
    \centering
    \resizebox{0.4\textwidth}{!}{
    \begin{tabular}{l|c}
        \toprule
        config & value \\
        \midrule
        training iteration & 20K  \\
        optimizer & Adam \\
        learning rate & $1\times10^{-5}$ \\
        batch size & 1024 \\
        weight decay & 0.001 \\
        optimizer momentum & $\beta_1,\beta_2$=0.9,0.999  \\
        data augmentation & \textit{RandomCropResize} \\
        \bottomrule
    \end{tabular}
    }
    \label{tab:pretrain-setting}
\end{table}

\subsection{Experimental setup for Figure~\ref{fig:ablation_248}}
\label{subsec:appen_248}
In Eq.~(\ref{equ:sum_r_def_delta}), we directly define the segment total rewards as the cosine similarity between the transition direction and the text embedding as follows:

\begin{equation}
    {\rm (Ours) \ }: \sum_t r_T(o_t,o_{t+1};\phi,\psi,l)=\mathcal{S}(\phi(o_{n+m})-\phi(o_{n}),\psi(l)),
\end{equation}

Like the potential-based reward for DeicisionNCE-P, this transition-direction reward bypasses all intermediary transitions and only focuses on the binary start-goal transition directions. This has been proven advantageous in action recognition~\citep{korbar2019scsampler}. However, we would like to provide a more detailed experimental evaluation to further support this.

In specific, we also define transition-direction rewards considering some intermediary transitions as follows:

\begin{equation}
    {\rm (4  \ frames) \ }: \sum_t r_T(o_t,o_{t+1};\phi,\psi,l)=\sum_{i=1}^4 \mathcal{S}(\phi(o_{n+\lfloor\frac{m\times{i}}{4}\rfloor})-\phi(o_{n+\lfloor\frac{m\times{(i-1)}}{4}\rfloor}),\psi(l)),
\end{equation}
\begin{equation}
    {\rm (8  \ frames) \ }: \sum_t r_T(o_t,o_{t+1};\phi,\psi,l)=\sum_{i=1}^8 \mathcal{S}(\phi(o_{n+\lfloor\frac{m\times{i}}{8} \rfloor})-\phi(o_{n+\lfloor\frac{m\times{(i-1)}}{8} \rfloor}),\psi(l)),
\end{equation}

This treatment will include more intermediary transitions when calculating the preference model. Then, we use these objectives to train representations and use the obtained representations to serve as input for downstream Frankakitchen LCBC tasks. The results are presented in Figure~\ref{fig:ablation_248}. Results clearly demonstrate that including more intermediary points may complicate the training process and hamper downstream policy learning. Therefore, we focus solely on the teleporting start-goal transitions to train representations.

\subsection{Experimental setup for Table~\ref{tab:cosine_sim_o1}}
\label{subsec:appen_table2}
To validate the effectiveness of DecisionNCE-P/T in clustering the first, potentially task-irrelevant image embeddings into similar positions, we randomly select 100 trajectories from the EPIC-KITCHEN-100 dataset. Subsequently, we calculate and compare the cosine similarities among the first image embeddings, denoted as $\phi(o_1)$, from these trajectories. Additionally, we assess the cosine similarities between $\phi(o_1)$ and $\psi(\Bar{l})$, where $\psi(\Bar{l})$ is derived by averaging all language embeddings within the EPIC-KITCHEN-100 annotations. The average result from both LIV and our DecisionNCE, acquired after five instances of random sampling, is presented in Table~\ref{tab:cosine_sim_o1}.

We chose LIV as the main baseline since it also builds on top of CLIP, which has the same representation embedding dimensions as our DecisionNCE.
We also calculate the $\phi(o_1)$ similarities for CLIP, and obtain a value of \textbf{0.99}. However, we must mention that this comparison is not meaningful in our setting since CLIP does not aim to align text and image at the task level. Moreover, the EPIC-KITCHEN-100 dataset is significantly smaller in scope compared to the dataset used for pre-training CLIP, making a direct comparison of values unfair. 
Additionally, we do not compare against R3M in this experiment as the vision embedding and language embedding for R3M have different dimensions and thus cannot calculate the distances or similarities between these two embeddings. In addition, R3M chose $\mathcal{S}$ as negative L2 distance, whereas DecisionNCE chose $\mathcal{S}$ as cosine similarity. Therefore, these two methods are not comparable in this setting due to the differences in their evaluation metrics.

\subsection{Environments}
\label{sub_app:env}
\textbf{Frankakichen}. Frankakitchen environment consists of a Franka robot interacting within a kitchen scene that contains diverse household kitchen objects such as microwaves and cabinets~\citep{gupta2019relay}. In our work, we follow the same 5 task definitions in LIV~\cite{r3m} and R3m~\citep{r3m}, reported in Table~\ref{tab:franka}. For each task, we have 1/3/5 demonstrations from~\citep{r3m}, thus the total dataset includes only 5/15/25 episodes, and each episode has 50 environmental steps. These small datasets are not enough to train a vision-language agent from scratch but becomes possible by pretraining the vision-language representations on abroad out-of-domain data in advance.

\begin{table}[h]
    \centering
    \caption{FrankaKitchen Tasks}
    \begin{tabular}{ll}
    \toprule
     Environment ID  & Language Instruction \\
     \midrule
     \texttt{kitchen\_micro\_open-v3} & \texttt{open microwave}\\
     \texttt{kitchen\_sdoor\_open-v3} & \texttt{slide cabinet}\\
     \texttt{kitchen\_ldoor\_open-v3} & \texttt{open left door}\\
     \texttt{kitchen\_knob1\_on-v3} & \texttt{turn on stove}\\
     \texttt{kitchen\_light\_on-v3} & \texttt{switch on light}\\
     \bottomrule
    \end{tabular}
    \label{tab:franka}
\end{table}

\textbf{WidowX RealRobot}. WidowX RealRobot environment contains a WidowX 6DoF robot arm executing diverse manipulation tasks. In specific, we evaluate on 5 distinct skills including \{\texttt{pick \& place, move, fold, flip, open \& close}\} comprising 9 sub-tasks as shown in Table~\ref{tab:real_robot}. For each task, we collect around 100 demonstrations using the demonstration collection system in BridgedataV2~\citep{walke2023bridgedata}. For each demonstration, the environmental steps are around 50 steps. We collect more data for WidowX RealRobot than Frankakitchen as the real robot environment is far more stochastic, where the object and robot locations are randomly initialized, and the scene also has lots of randomly located distractors with varied shape and color.

\begin{table}[h]
    \centering
    \caption{WidowX RealRobot Tasks}
    \begin{tabular}{ll}
    \toprule
     Environment ID  & Language Instruction \\
     \midrule
     \texttt{Red cup on silver pan} & \texttt{Pick up the red cup and place it on the silver pan}\\
     \texttt{Red cup on red plate} & \texttt{Pick up the red cup and place it on the red plate}\\
     \texttt{Duck on green plate} & \texttt{Pick up the duck and place it on the green plate}\\
     \texttt{Duck in pot} & \texttt{Pick up the duck and place it in the pot}\\
     \texttt{Move pot} & \texttt{Move the pot from right to left}\\
     \texttt{Fold cloth} & \texttt{Fold the cloth from right to left}\\
     \texttt{Flip the red cup upright} & \texttt{Flip the red cup upright}\\
     \texttt{Open the microwave} & \texttt{Open the microwave}\\
     \texttt{Close the microwave} & \texttt{Close the microwave}\\
     \bottomrule
    \end{tabular}
    \label{tab:real_robot}
\end{table}

\subsection{LCBC Experiment Details}
Here, we train a Language-Conditioned Behavior Cloning (LCBC) policy take the frozen vision-language representations as inputs. In details, the raw images $o$ and language instructions $l$ will be fed into the learned vision-language encoder and obtain the representations $\phi(o)$ and $\psi(l)$. Then, the policy $\pi({\rm StopGrad}(\phi(o)), {\rm StopGrad}(\psi(l)))$ take these representations as input to fit the actions in demonstrations. Note that we freeze the pretrained representations and only optimize a small MLP to avoid overfit, following~\citep{liv, vip, r3m, karamcheti2023language}, as the downstream domain-specific data is quite small. The policy input also includes the proprioception information, which is concatenated with the frozen vision-language representation and fed into the downstream MLP.

\begin{figure}[h]
    \centering
    \includegraphics[width=0.9\textwidth]{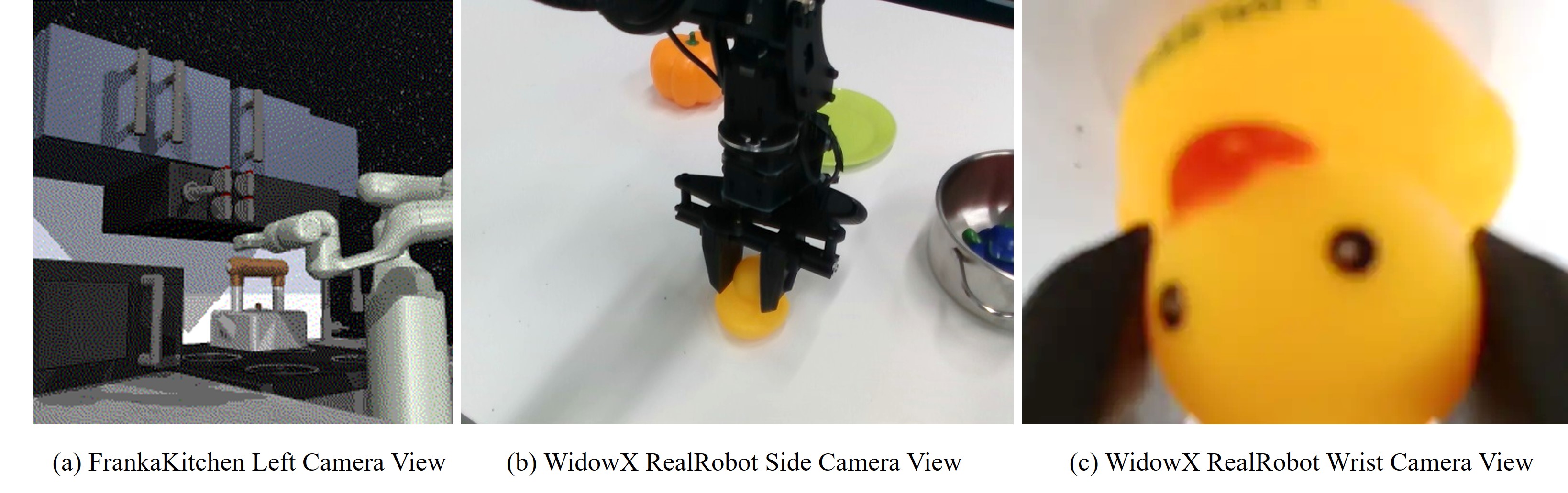}
    \caption{Visual input view for LCBC policy.}
    \label{fig:view}
\end{figure}

\textbf{FrankaKitchen}. For FrankaKitchen experiments, we take single left camera view as the vision input, as shown in Figure~\ref{fig:view} (a). We choose a small learning rate 1e-4, 16 batch size and 2e4 gradient steps, as the simulation dataset is quite small. Detailed hyper-parameters can be found in Table~\ref{tab:lcbc_details}. We evaluate the policy for 25 episodes per 2e3 gradient steps and report the max success rate over the training following previous works~\citep{liv, r3m, vip}.

\textbf{WidowX RealRobot}. For real robot experiments, we take both a side camera view and a wrist camera view as the vision inputs, as shown in Figure~\ref{fig:view} (b)-(c). We choose a relatively large learning rate 1e-3, 64 batch size, as the real robot dataset contains more data. We also use data augmentation on the brightness to enhance the generalizability. See Table~\ref{tab:lcbc_details} for more details. We evaluate the success rate only for the last checkpoint after the training, as the real robot evaluation is extremely time-costly. We rollout 10 episodes for each checkpoint, and evaluate 3 policy checkpoints trained with different seeds for each pretrained representation.

\begin{table}[h]
    \centering
    \caption{LCBC experiment details}
    \begin{tabular}{llll}
    \toprule
     ~ & FrankaKitchen & WidowX RealRobot \\
    \midrule
    MLP Architecture & [256, 256] & [256, 256]\\
    Activation Function & ReLU & ReLU \\
    Optimizer & Adam & Adam \\
    Learning Rate & 1e-4 & 1e-3 \\
    Batch Size & 16 & 64 \\
    Gradient Steps & 2e4 & 200 epoch \\
    Proprioception & Yes & Yes \\
    Augmentation & No & Yes \\
    \bottomrule
    \end{tabular}
    \label{tab:lcbc_details}
\end{table}

\textbf{Baseline Setups}. We use the official released checkpoints of the vision/vision-language encoder from previous works. For VIP~\citep{vip} baseline, we use the pretrained DistilBERT~\citep{sanh2019distilbert} as the language encoder, which is the same language encoder used in R3M~\citep{r3m}, as VIP studies only the visual modality. For baselines including R3M~\citep{r3m}, LIV~\citep{liv}, VIP~\citep{vip} and CLIP~\citep{clip}, we adhere the same hyper-parameters in Table~\ref{tab:lcbc_details} to train the downstream LCBC policy to ensure fair comparisons.

\subsection{Frankakichen Planning Experiments}
\label{frankakitchen}
In this section, we'll describe more experimental details of language-reward planning. As we can directly roll out trajectories in FrankaKitchen environment, we combine ground-truth environment dynamics with reward induced by learned representation as the model in MPPI. This also aligns with the setting in \citep{vip} and \citep{liv}. For CLIP, LIV, DecisionNCE-P and DecisionNCE-T, we all use $\mathcal{S}(\phi(o_{n+1}), \psi(l)) - \mathcal{S}(\phi(o_{n}), \psi(l))$ as reward. For R3M, the model contains a reward model $R(o_0, o_n, l)$ whose inputs are initial observation, present observation and language goal and output is a score estimating the task completion degree. So we take $R(o_0, o_{n+1}, l) - R(o_0, o_n, l)$ as reward for R3M. For all reward model we take discount factor $\gamma = 1.0$ to calculate return.

As is pointed out in \citep{liv}, tasks in FrankaKitchen demand a high level of exploration. So we follow the same setting and warmstart the action search with a fixed open-loop sequence came from expert demonstration. The warmstart stage won't directly solve the task, but will bring the robot end-effector to the vicinity of the task objective. This reduces the number of planning horizon and proposed action sequences and speed up the planning stage.

We list all hyperparameters of MPPI in Table \ref{tab:mppi_hyperparam}. It's worth mention that because $\mathcal{S}(\phi(o_{n}), \psi(l))$ could be the cosine similarity between image and language embedding, $\mathcal{S}(\phi(o_{n+1}), \psi(l)) - \mathcal{S}(\phi(o_{n}), \psi(l))$ lies in $[-2, 2]$ and is usually small in this case. We'll then normalize the return among all proposed action sequences through $R_{normed} = \frac{R - R_{mean}}{R_{std}}$.

\begin{table}[h]
\small
\setlength{\tabcolsep}{2pt}
    \centering
    \caption{MPPI Hyperparameters (parameters not listed are set to default values in \citep{liv})}
    \begin{tabular}{ll}
    \toprule
      & FrankaKitchen\\
     \midrule
     Planner & MPPI~\citep{williams2017model}\\
     Planning Horizon & 50\\
     Proposed Action Sequences & 64\\
     Optimization Iteration & 1\\
     Temperature & 10.0\\
     \bottomrule
    \end{tabular}
    \label{tab:mppi_hyperparam}
\end{table}

\section{Limitations \& Discussions \& Future Work}
\label{appen:limitation}
Here, we discuss our limitations, potential solutions to our limitations and interesting future works.
\begin{enumerate}
    \item \textbf{Implicit Preference}. Note some few counterexample also exist that may violate the \textit{implicit preference} assumption, which will bring noisy preference labels. For instance, consider two videos, one for ``open microwave" and the other for ``open the microwave". In principle, these two videos should not be treated as different, as they complete the same task. However, in our paper, these videos are treated as suboptimal for the mismatched language instructions as the instructions are a little bit different. 
    
    \textit{Solution and future work:} This problem can be solved by proper data filtering and reorganization but might demand tremendous human costs. In this paper, we directly utilize the original dataset $\mathcal{D}$ for convenience, but observe good robustness to these noisy samples, which have also been observed in previous studies~\citep{nair2022learning}. We believe this is an interesting observation and leave this direction as a future work.

    \item  \textbf{Reward definition}. In this current version, we only evaluate two types of reward definitions in the BT model. Thus, we cannot guarantee our design choices must be the optimal solution. 

    \textit{Solution and future work:} We have shown that by simply designing the rewards as potential-based or transition-direction rewards in embedding space, we can inherently recover an implicit time contrastive learning, naturally marries language grounding and temporal consistency in an elegant and unified objective. Therefore, we believe there must be lots of efficient reward forms still under-explored and encourage the researchers to investigate this.


    
    \item \textbf{Out-of-domain Dataset}. In this current version, we only use the EPIC-KITCHEN-100 dataset~\citep{damen2018scaling} to pretrain the representations. Although this is the common choice of previous works~\citep{liv}, we observe the language diversity in this dataset is far more limited, including lots of simple instructions such as \texttt{put \& take \& open \& close \& move something}. The reason is straightforward, as language is highly abstract, which can describe many videos with the same language instruction. Therefore, the language diversity is quite limited, which may be not appropriate to train a generalizable language encoder. 
    
    \textit{Solution and future work:} To address this, one straightforward solution is to include more out-of-domain data, including but not limited to Ego4D~\citep{grauman2022ego4d}, Something-Something~\citep{goyal2017something}, and Open-X-embodiment dataset~\citep{open_x_embodiment_rt_x_2023}. However, we find the language diversity in these datasets is still limited compared to image diversity. To further tackle this challenge, one promising direction is to utilize the powerful Large Language Models (LLMs)~\citep{touvron2023llama, wang2023openchat} to augment the language instruction reasonably.
\end{enumerate}

Overall, although some design choices seem quite simple and some potential limitations exist, they smartly tackle all challenges faced by previous studies.  Moreover, in this work, we have demonstrated the superior effectiveness and versatility of DecisionNCE to extract value features from out-of-domain data to facilitate efficient downstream policy learning and provide several solutions to tackle these limitations. Therefore, we leave these further detailed improvements for future work. In summary, DecisionNCE provides an elegant and principled solution for decision-centric vision-language representation and reward learning. 
\end{document}